\documentclass[5p,times,sort&compress]{elsarticle}
\usepackage{xcolor, soul}
\usepackage[bottom]{footmisc}
\usepackage{fleqn,multirow,graphicx,wrapfig,lineno,hyperref}
\usepackage[nodots]{numcompress}
\usepackage{comment}
\hypersetup{colorlinks=true}
\modulolinenumbers[5]
\journal{Mechatronics}
\graphicspath{ {figs/} }
\sethlcolor{yellow}

\definecolor{bronze}{rgb}{0.8, 0.5, 0.2}
\definecolor{blue}{rgb}{0, 0, 1}
\definecolor{green}{rgb}{0, 1, 0}
\definecolor{black}{rgb}{0,0,0}
\newcommand{\revisedP}[1]{{\color{black}#1}}
\newcommand{\revisedS}[1]{{\color{black}#1}}

\newcommand{\revisedSecondRevision}[1]{{\color{black}#1}}
\renewcommand{\hl}{\revisedP}

\begin{document}

\begin{frontmatter}

\title{An adaptive admittance controller for collaborative drilling with a robot based on subtask classification via deep learning}

\author[mymainaddress,mysecondaryaddress]{Berk Guler}
\author[mymainaddress,mysecondaryaddress]{Pouya P. Niaz}
\author[mymainaddress,mysecondaryaddress]{Alireza Madani}
\author[mythirdaddress]{Yusuf Aydin}

\author[mymainaddress,mysecondaryaddress]{Cagatay Basdogan\corref{mycorrespondingauthor}}

\cortext[mycorrespondingauthor]{Corresponding author}
\ead{cbasdogan@ku.edu.tr}

\address[mymainaddress]{Robotics and Mechatronics Laboratory, Koc University, 34450, Istanbul, Turkey}
\address[mysecondaryaddress]{KUIS AI Center, Koc University, 34450, Istanbul, Turkey}
\address[mythirdaddress]{Faculty of Engineering, MEF University, 34396, Istanbul, Turkey}

\begin{abstract}
\revisedP{In this paper, we propose a supervised learning approach based on an Artificial Neural Network (ANN) model for real-time classification of subtasks in a physical human-robot interaction (pHRI) task involving contact with a stiff environment. In this regard, we consider three subtasks for a given pHRI task: \textit{Idle}, \textit{Driving}, and \textit{Contact}. Based on this classification, the parameters of an admittance controller that regulates the interaction between human and robot are adjusted adaptively in real time to make the robot more transparent to the operator (i.e. less resistant) during the \textit{Driving} phase and more stable during the \textit{Contact} phase. The \textit{Idle} phase is primarily used to detect the initiation of task. Experimental results have shown that the ANN model can learn to detect the subtasks under different admittance controller conditions with an accuracy of 98\% for 12 participants. Finally, we show that the admittance adaptation based on the proposed subtask classifier leads to 20\% lower human effort (i.e. higher transparency) in the \textit{Driving} phase and 25\% lower oscillation amplitude (i.e. higher stability) during drilling in the \textit{Contact} phase compared to an admittance controller with fixed parameters.}
\end{abstract}


\end{frontmatter}

\section{Introduction}
	\label{sec:Intro}
The application of robots in the manufacturing industry has grown considerably during the \revisedSecondRevision{last three decades} due to the improvements in precision and accuracy of robots, as well as the fact that robots have become easier to program and deploy at factories \cite{Ben-Ari2017,Heyer2010}. However, manual labor, which creates many ergonomic and health problems for human workers, has not been completely eradicated, as many repetitive small-batch processes still require human-level awareness and decision making.

\begin{figure}[t!]
	\includegraphics[width=\columnwidth]{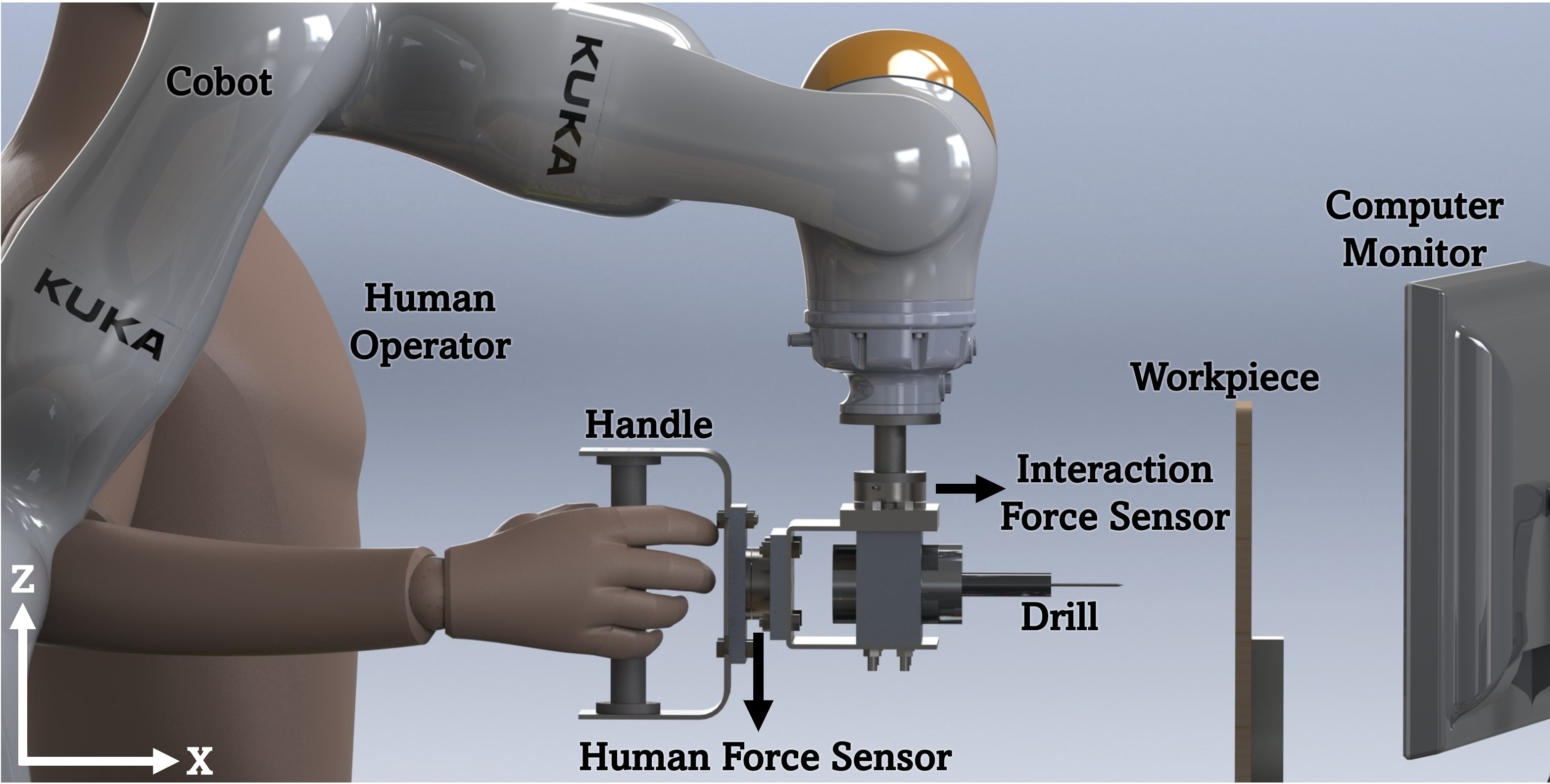}
	\centering
	\caption{Components of our hardware setup for collaborative drilling with a robot.}
	\label{fig:cad}
\end{figure}
In recent years, the mechanical adeptness of robots, compounded by human intelligence, has lead to the emergence of physical human-robot interaction (pHRI), the idea of robots and humans collaborating to perform many tasks more efficiently in such manufacturing environments \cite{Ajoudani2017}. This has been made easier by the emergence of machine/deep learning methods that can help robots learn to work with human operators \cite{Mukherjee2022}. Such collaboration can provide human operators with precise but also compliant tools to perform their tasks with reduced health risk and better ergonomics  \cite{Villani2018,Cherubini2016a}. In the foreseeable future, the human workforce is anticipated to cooperate with robotic systems rather than be replaced by them entirely \cite{Haddadin2016}. This can already be noticed by observing recent market reports of collaborative robots (cobots) \cite{MarketsAndMarkets2021}. 

Thanks to the current progress in machine/deep learning, the intention of the human operator in pHRI can be predicted up to some extent. This enables the collaborative robot to comply with the human by adapting the interaction controller, further reducing human effort and maximizing task efficiency \cite{Losey2018}. Selvaggio et al. \cite{Selvaggio2021} provide a survey of how different methods have been used for adaptive control in pHRI, through the concepts of shared control and shared autonomy, in order to maximize task efficiency in pHRI.

Admittance control has become a popular method for pHRI, as it offers simple means of reading the force input and outputting a reference velocity in task-space which the robot can follow
\cite{Duchaine2007,duchaine2009safe}. \revisedP{Since force sensors are readily available in collaborative robots such as ours, admittance control is a viable option for physical interaction with robots that are motion controlled}\cite{AUNG2017110}. 



\revisedP{In order to reduce human effort and increase task efficiency, adaptive admittance control is often utilized in pHRI in which admittance mass and damping are adjusted in real time based on human intention. There are mainly two different approaches for detecting human intention in pHRI: a) rule-based, and b) learning-based. Rule-based approaches have been frequently used to adapt an interaction controller based on recognized human intention to take leader or follower role, or to accelerate or decelerate a co-manipulated object \cite{Oguz2012,Kucukyilmaz2013,Mortl2012,Medina2013,Ikeura1995,Duchaine2007,Aydin2014,Hamad2021,Campeau-Lecours2016,Sirintuna2020}. In such systems, a predefined and fixed or manually tuned set of heuristics is implemented for detecting human intention in real time.}

\revisedP{Albeit fast and easy to implement, rule-based approaches for detecting human intention tend to have adequate performance under certain circumstances only and cannot easily be generalized to different pHRI tasks of similar nature. This is because rule-based approaches often use task-specific variables such as position of robot's end-effector \cite{Sirintuna2020}, or \revisedSecondRevision{extremum} values of force, velocity or their derivative to detect human intention \cite{Mortl2012,Medina2013,Ikeura1995,Duchaine2007,Aydin2014,Hamad2021}. The range of such variables are task and environment-dependent. As a result, such systems need to be manually tuned when the task or environment changes.} 

\revisedP{Alternatively, researchers have utilized machine/deep learning approaches more recently to detect human intention, so as to grant intelligence to robots in pHRI systems}. Machine learning methods can use kinematic, kinetic, audiovisual and other sensory data, to detect human intention and enable to adapt the interaction controller accordingly \cite{Li2020,Mazhar2019,Dong2019,Liu2018a,Delpreto2019,Grafakos2017,Wu2019,Sirintuna2020a,Ge2011,Li2014,Rozo2016,Townsend2017,Sharkawy2020,Dimeas2015,Wu2020,Ghadirzadeh2016,Buchli2011,Du2017}. 

In this paper, we argue that most pHRI tasks in small-batch manufacturing operations can be divided into multiple subtasks, which can be detected by machine/deep learning techniques. Oftentimes, the human intention itself depends on which subtask is being performed at any stage as argued in \cite{utkuThesis}. Accordingly, different subtasks require different control parameters. For instance, when a human operator is pushing and guiding a robot in free space, more transparent interaction (\revisedP{i.e. less resistance to human motion}) is desired to reduce human effort, which often corresponds to utilizing low damping values in the admittance controller. On the other hand, when the robot, for example, is being used to drill a hole on a workpiece, it is in contact with a stiff environment, and therefore a more stable interaction is required. This usually corresponds to utilizing high damping values in the admittance controller. Consequently, if a pHRI system can understand which subtask the human operator is executing at any given moment, it can adapt the control parameters of the robot (admittance damping in the above discussion) accordingly. This can lead to an efficient execution of the task while the trade-off between transparency and stability is balanced well.

We consider a collaborative drilling scenario (Fig. \ref{fig:cad}) as our representative pHRI task to test our idea. The drilling task is divided into three subtasks (see Fig. \ref{fig:subtasks}): \textit{Subtask 1: Idle} at the beginning before the human grabs the handle, \textit{Subtask 2: Driving} for when the robot is guided by the human in free space towards the workpiece, and \textit{Subtask 3: Contact} for when the robot is in contact with the workpiece, and drilling process is underway. We use artificial neural networks (ANNs) to detect the subtasks in real-time. The predicted subtask (\textit{Idle}, \textit{Driving}, or \textit{Contact}) at any instant is then passed to an adaptive control policy, which simply sets the desired parameters of an admittance controller accordingly. The transition from one subtask to another one is handled by a linear interpolation between the corresponding values of the parameters.

\subsection{Related work}

\revisedP{We review the earlier studies on pHRI in two groups based on how human intention is detected for the adaptation of an interaction controller: the ones using a) rule-based and b) learning-based approaches.}

\subsubsection{Rule-based approaches}

A number of studies have utilized heuristics based on kinematic and/or kinetic data recorded during a pHRI task in order to regulate the interaction between the human and the robot and maximize the task efficiency. For instance, Oguz et al. \cite{Oguz2012} and Kucukyilmaz et al. \cite{Kucukyilmaz2013} developed methods for role exchange in human-robot interaction based on haptic cues including interaction forces. Their method was adopted by Mortl et al. \cite{Mortl2012} to adjust the contribution of a humanoid robot to a collaborative table transportation task. Likewise, Medina et al. \cite{Medina2013} utilized human force in a rule-based mechanism to adjust the robot's degree of contribution to a task. Ikeura and Inooka \cite{Ikeura1995} developed a rule-based approach based on Cartesian velocity for adapting the damping parameter of an admittance controller. Duchaine et al. \cite{Duchaine2007} utilized both velocity and derivative of interaction force to estimate the state of human intention as acceleration or deceleration and then adjust the admittance damping accordingly. Aydin et al. \cite{Aydin2014} improved this idea by adding a fuzzy intention estimator for smoother transitioning between those states, which was later further improved and utilized by Hamad et al. to adjust the gain of an admittance controller in a collaborative object transportation task \cite{Hamad2021}. As another example, Campeau-Lecours et al. \cite{Campeau-Lecours2016} used predefined thresholds of a so-called ``Vibration Index", which is calculated online from measured velocity, to tune the gains of an adaptive admittance controller to minimize the vibrations during contact interactions with stiff environments.

In pHRI tasks which involve contact with stiff environments, such as drilling a workpiece with the help of a robot, stable interaction during contact is critical for the safety of the human operator. In our earlier studies, Sirintuna et al. \cite{Sirintuna2020} utilized the position of the robot's end-effector to adapt an admittance controller in a collaborative drilling task. Assuming the position of the workpiece is known in advance, a rule-based adaptation method was implemented such that the value of admittance damping is well within stable bounds when the drilling starts.

As mentioned earlier, rule-based approaches for detecting human intention in pHRI tasks typically utilize kinematic and/or kinetic data. Such variables are highly task-specific, and their values depends on the characteristics of the user and the environment even for the same task. Hence, rule-based approaches need to be manually tuned whenever such changes occur. In contrary, methods utilizing machine/deep learning often have the advantage of flexibility and partial task independence. Such methods typically do not directly depend on the thresholds of kinematic and kinetic variables. For example, in a collaborative drilling task, the thresholds for interaction or human force heavily depend on the user and the material properties of the workpiece being drilled, thereby making rule-based approaches almost impractical for real-life applications.

\subsubsection{Learning-based approaches}

As a more promising approach for improving the efficiency in pHRI, machine/deep learning methods have gained a considerable attention recently, due to their flexibility, portability, and task independence. Significant amount of research has been conducted on using supervised learning methods for adaptive control in pHRI. 
Several studies have utilized visual and auditory information such as images, videos, and voice signals to detect human intention via a learning model \cite{Li2020,Mazhar2019,Dong2019}. For example, Liu et al. \cite{Liu2018a} trained a deep learning model utilizing body posture, hand gesture, and voice command together to adapt an interaction controller. The problem with such systems is that they require external cameras and motion capture systems for implementation. Furthermore, a large number of images needs to be captured and processed for training the learning model.

As an alternative to audiovisual signals in detecting human intention, some researchers have chosen surface electromyography (sEMG) signals, arguing that it directly shows the activation levels of muscles in human limbs, thereby providing a more direct interface of predicting what the human operator is trying to do. Delpreto and Rus \cite{Delpreto2019} used EMG signals for tuning an adaptive controller in a robot-assisted lifting task. Grafakos et al. \cite{Grafakos2017} proposed a variable admittance control scheme utilizing EMG signals as input to minimize human effort. Wu et al. \cite{Wu2019} developed a variable impedance controller utilizing EMG signals to teach a robot how to perform a collaborative task via the Learning from Demonstration (LfD) method. Sirintuna et al. \cite{Sirintuna2020a} trained an ANN model using EMG signals to predict the direction of movement intended by the human, and then adapted an admittance controller to confine the human motion to the predicted direction. The fundamental shortcoming of using EMG signals for human intention prediction is that they require additional equipment attached to human limbs, reducing their practicality in pHRI tasks performed in manufacturing environments. EMG data also requires additional preprocessing since the ranges of EMG signals may vary across the trials of a user and also among the different users.

Some other researchers have opted to simply use kinematic data such as end-effector velocity or kinetic data such as interaction force, to train supervised learning models. Such methods are more practical since velocity can be easily obtained by differentiating the position and force/torque sensors are typically available in collaborative robots nowadays. Sam Ge et al. \cite{Ge2011} developed an Artificial Neural Network (ANN) model to recognize human intention using kinematic data, including end-effector velocity. Li et al. \cite{Li2014} utilized ANNs in a similar manner to estimate the human arm impedance and the desired trajectory of the human operator. Rozo et al. \cite{Rozo2016} utilized end-effector position and interaction force together to train a machine learning algorithm for LfD to estimate human arm stiffness and desired movement trajectories in cooperative transportation and assembly tasks. Sharkawy et al. \cite{Sharkawy2020} trained an ANN model online for a collaborative manipulation task and adjusted the admittance mass of the controller to obtain a manipulation trajectory that is close to the minimum-jerk trajectory.

Some researchers have argued the importance of learning interaction patterns from human dyads to transfer this knowledge to human-robot collaboration. Madan et al. \cite{madan2014recognition} used support vector machine (SVM) classifier to successfully learn three interaction types from the velocity and force data of collaborating human dyads: 1) work in harmony, 2) cope with conflicts, or 3) remain passive during interaction. Al-Saadi et al. \cite{al2020novel} showed that a set of features derived from force data alone is sufficient for the successful classification of interactive motor behaviors encountered in human dyads during collaborative object manipulation. Townsend et al. \cite{Townsend2017} used velocity and acceleration signals acquired from human dyads in co-manipulation tasks to train an ANN model that predicts velocity values for a few time steps into the future. They then deployed that model to adjust an adaptive impedance controller in human-robot collaboration to perform the same task. 

Thanks in large part to the unpredictable and highly nonlinear movement behavior of the human operator under different conditions, reinforcement learning (RL) is also a popular choice of machine learning method for adaptive control or intention recognition in pHRI. Dimeas and Asparagathos \cite{Dimeas2015} proposed to utilize minimum jerk as a reward function for a reinforcement learning method with the objective of altering the admittance damping in an optimum manner. This method may not work well in pHRI tasks for manufacturing since they typically involve contact with a stiff environment and hence rapid change in the closed-loop dynamics, which results in a jerky response in the velocity and force profiles. Wu et al. \cite{Wu2020} used RL, specifically Q-Learning, to optimize an adaptive impedance controller for a co-manipulation task. They aimed to achieve a desired motion trajectory while minimizing the internal forces acted on the manipulated object and the total energy consumption. Ghadirzadeh et al. \cite{Ghadirzadeh2016} developed a Q-learning approach so that a user collaborating with a robot can keep a ball rolling on a beam at a desired target location with minimum effort. They trained a Gaussian process regressor as the Q-function to estimate the outcome of state-action pairs instead of interacting with the robot directly. They then selected policies in adjusting the impedance of the controller such that the interaction force is minimized during the task. Buchli et al. \cite{Buchli2011} used a PI\textsuperscript{2} method (Policy Improvement with Path Integrals) for optimizing an adaptive impedance control via gain scheduling to minimize the control effort. Du et al. \cite{Du2017} utilized the fuzzy Sarsa($\lambda$)-Learning method for minimizing the jerk in human hand during a minimally invasive robotic surgery task via adaptive admittance control in joint space, taking inputs from joint external torques, and combining them with joint velocities and accelerations as state variables.

\subsection{Contributions}
One of the shortcomings of the studies cited above is that they consider the pHRI task as a whole and do not focus on its phases (subtasks). However, most small-batch pHRI tasks in manufacturing settings, such as drilling, polishing, cutting, sanding, welding and soldering, fastening, etc., involve multiple phases (subtasks) which require different control parameters. For example, during the so-called \textit{Driving} phase (subtask 2) of a collaborative drilling task, the user brings the drill close to the workpiece by manually guiding the robot in free space. A compliant behavior (minimal resistance to the user) is expected from the robot at this phase. However, during the \textit{Contact} phase (subtask 3), the drill is in contact with the workpiece and a more rigid behavior and stable operation is desired. These two phases with conflicting natures require different control parameters to be used with the interaction controller to render a more effective collaboration.

\revisedSecondRevision{
There are a limited number of studies in related literature, in which the type of a task executed by humans is recognized. For instance, ``enriched semantic event chains" have been used for detecting the type of object manipulation actions performed by a human, as in ``above", ``around", ``below", ``getting close", ``moving apart", etc. from visual data~\cite{ZIAEETABAR2018}. Similarly, Borras et al. showed that the manipulation of textile objects requires sequentially executed manipulation primitives and some of such primitives can be performed better with different grasp types~\cite{Borras2020}. However, to the best of the authors' knowledge, no earlier pHRI study involving contact with an environment has adapted an interaction controller based on subtask recognition.} In this study, we propose a new approach for classification of the subtasks of a pHRI task using ANNs to adjust the parameters of an admittance controller accordingly for more effective collaboration between human and robot.

The contributions of the present study to the current state of the art are as follows:
\begin{itemize}
\item We propose an adaptive admittance controller whose parameters are adjusted based on the subtasks of a pHRI task. The potential benefits of such an approach has been already tested by our group for an abstract pHRI study, in which the environment was represented by a mechanical spring \cite{utkuThesis}. In this study, however, the proposed approach has been tested on a real-life scenario. Collaborative drilling is used as a representative pHRI task that involves contact with a stiff environment, and consists of phases with differing interaction dynamics and control requirements.

\item We argue that a learning-based approach
is more effective than a rule-based one to detect subtasks of a pHRI task. Rule-based approaches requires fine-tuning of the rules for different subtasks and environmental conditions and may not be possible to implement for complex pHRI tasks. On the other hand, a learning-based approach is shown to be more robust to the uncertainties mentioned above given a diverse dataset for training. In our study, ANNs are used for training a classifier that detects subtasks in real time during the execution of a collaborative drilling task. The control parameters of an admittance controller are altered in real-time based on the detected subtasks for enabling more effective collaboration.

\item Our earlier studies have already shown the benefits of fractional order admittance controller over the integer order one especially for the contact phase of a pHRI task \cite{Aydin2018a,Sirintuna2020}. In this study, we further show that adapting not only the damping but also the fractional order of an admittance controller based on the subtasks detected by the classifier leads to a more effective pHRI in terms of contact stability. 
\end{itemize}

The rest of the present paper is organized as follows:\\

\hyperref[sec:Approach]{Section 2} explains the technical approach followed in the study, including the hardware setup, the closed-loop control system utilizing the admittance controller, its adaptation policy, the deep learning method used for subtask detection and how it is trained. \hyperref[sec:Experiments]{Section 3} introduces our pHRI experiments in full detail, including the experimental protocol and conditions, information about the participants, and the methods used for data collection and processing. \hyperref[sec:Results]{Section 4} reports the results of experimental data analyses. It explains the performance metrics used for comparison and contrast. It reports the performance results of the subtask classifier ANN model under the fixed and adaptive admittance controllers.
\hyperref[sec:Discussion]{Section 5} offers insights into the results obtained by the study and the details for the successful implementation of the proposed approach. \hyperref[sec:Conclusion]{Section 6} provides conclusive remarks, as well as possible future directions.

\begin{figure*}[t!]
	\includegraphics[width=0.8\textwidth]{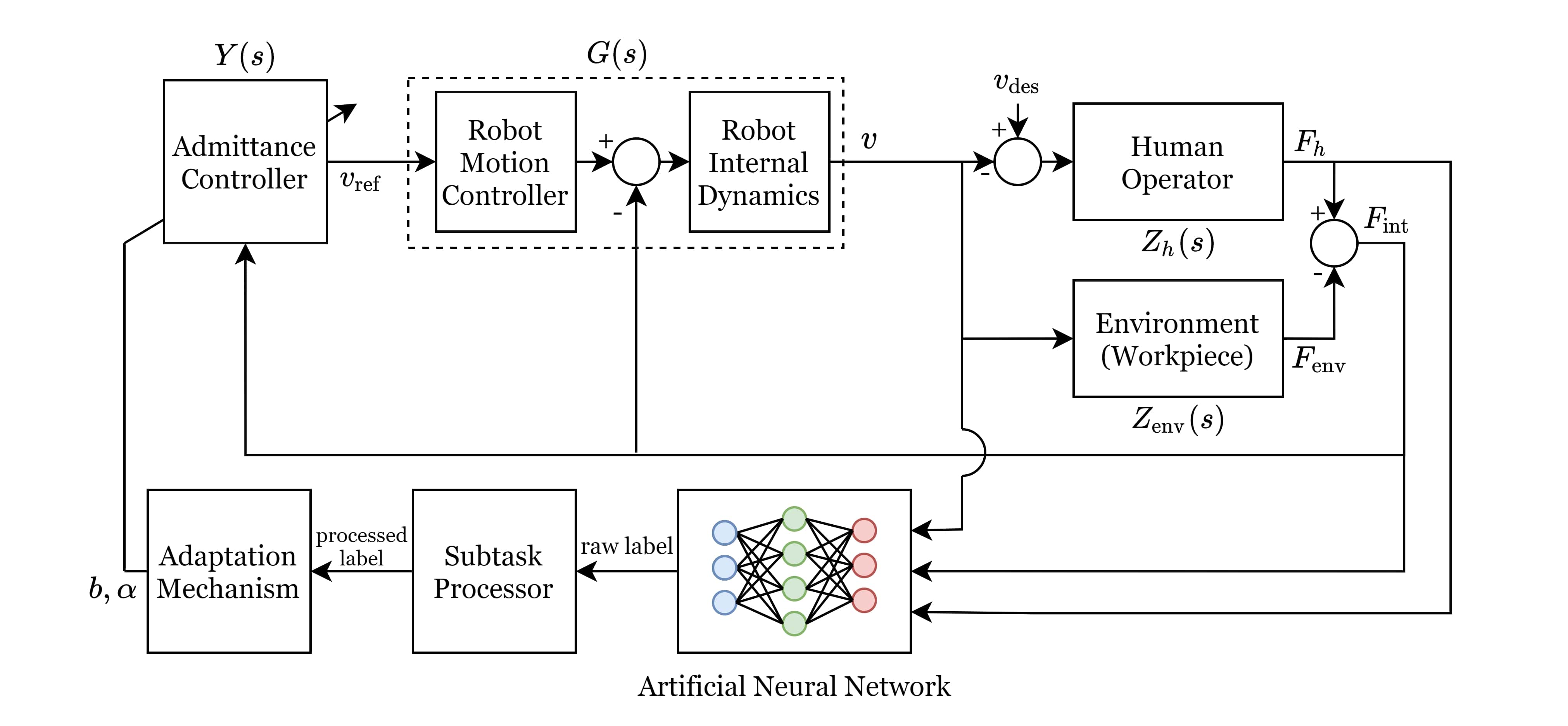}
	\centering
	\caption{Closed-loop control system used in our pHRI study.}
	\label{fig:control}
\end{figure*}

\section{Approach}
	\label{sec:Approach}

The approach followed in this study is to divide a pHRI task into subtasks, and using a deep learning model to estimate those subtasks in real-time for successful adaptation of the controller parameters. The pHRI task selected for the study is collaborative drilling which demands a trade-off between transparency and stability. The nature of trade-off was investigated in depth in our earlier studies \cite{Aydin2018a,Aydin2020}. Our objective in adaptation is to adjust the parameters of the admittance controller such that the robot is transparent to the user and hence human effort is minimized during \textit{Driving} phase (subtask 2) while stability is improved during \textit{Contact} phase (subtask 3).

The learning model takes robot's \textit{interaction force}, the \textit{force exerted by human operator} alongside with the \textit{velocity}, with which the human drives the robot, as the time-series input at each instant of the task execution and outputs the subtask that the user is currently executing. The  parameters of the admittance controller are set according to the estimated subtask and the transition between the different subtasks is handled by linear interpolation between the values of control parameters used for the previous and the current phases.

\subsection{Hardware Setup}
The hardware setup used in this study is illustrated in Fig.~\ref{fig:cad}. It is composed of the following major components:
\begin{itemize}
\item Robot: a 7R, KUKA LBR iiwa 7 R800 cobot.
\item Drill: Includes a DC motor operable between 0 and 48 volts and a drill bit attached to the motor.
\item Force Sensors: Two ATI Mini45 force/torque sensors are employed. One is used for measuring the interaction force, placed between the robot end-effector and the drill, and the other is used for measuring the human force, placed between the handle and the drill.
\end{itemize}

\begin{figure*}[t!]
	\includegraphics[width=\textwidth]{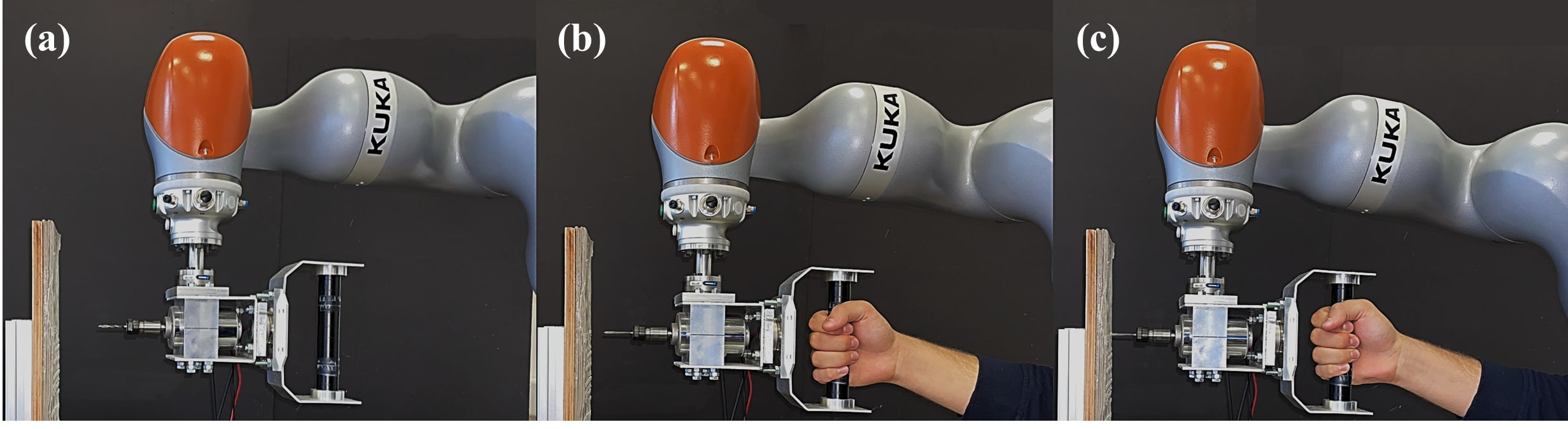}
	\centering
	\caption{Subtasks of the pHRI task; (a) \textit{Idle}: No contact with the handle yet; (b) \textit{Driving}: The operator is guiding the drill attached to the robot in free space; (c) \textit{Contact}: The operator is drilling the workpiece}
	\label{fig:subtasks}
\end{figure*}

\subsection{Closed-loop control system}
The schematics of the control system utilized in our pHRI scenario can be seen in Fig. \ref{fig:control}. According to this figure, the human, $F_{h}$, and environment, $F_{\mathrm{env}}$, both act on the end-effector, applying force to it. The resultant force, $F_{\mathrm{int}}$, is measured by the interaction force sensor and fed to the admittance controller. The admittance controller calculates the reference velocity for the robot, $v_{\mathrm{ref}}$ to follow while the actual velocity achieved by the robot using its internal motion controller is $v$. \revisedS{Finally, $v_\mathrm{des}$ represents the desired velocity at which the human operator intends to move the end-effector of the robot. }

Two types of admittance controllers are used and discussed in the present paper, in order to demonstrate the flexibility of our subtask classifier to different kinds of control schemes.

\paragraph{\textbf{Integer-Order Admittance Controller (IOAC)}} In Fig \ref{fig:control}, $Y(s)$ is the transfer function of the admittance controller. For a typical integer-order admittance controller (IOAC), this transfer function takes the form:
\begin{equation}
    Y(s) = \frac{V_{\mathrm{ref}}(s)}{F_{\mathrm{int}}(s)} = \frac{1}{ms+b}
\end{equation}
where $V_{\mathrm{ref}}(s)\,\mathrm{[m/s]}$ is the reference velocity generated by the admittance controller for the robot to follow, $F_{\mathrm{int}}(s)\,\mathrm{[N]}$ is the input interaction force along the direction of motion (positive $X$ direction in our implementation, see Fig. \ref{fig:cad}), $m\,\mathrm{[Kg]}$ is the admittance mass parameter, $b\,\mathrm{[Ns/m]}$ is the admittance damping parameter, and $s$ is the Laplace variable.

\paragraph{\textbf{Fractional-Order Admittance Controller (FOAC)}} Aydin et al. \cite{Aydin2018a} proposed a fractional-order admittance controller (FOAC) for pHRI, and showed that it offered a better trade-off between transparency to human intended motion and stability robustness. This controller was also used by Sirintuna et al. \cite{Sirintuna2020} in drilling experiments nearly identical to that of ours, and the authors reached a similar conclusion. For the FOAC proposed in the above references, the admittance control transfer function is as follows:
\begin{equation}
    Y(s) = \frac{V_{\mathrm{ref}}(s)}{F_{\mathrm{int}}(s)} = \frac{1}{ms^\alpha+b}
    \label{eq:eq2}
\end{equation}
where $\alpha$ is the fractional order, varying between $0 \leq \alpha \leq 1$. \revisedS{More information can be found in \cite{Aydin2018a, Sirintuna2020} regarding the details of fractional order interaction controllers and their dynamics.}

For the closed-loop control system shown in Fig. \ref{fig:control}, $G(s) = V(s)/V_{\mathrm{ref}}(s)$ is the transfer function model for the internal dynamics and controller of the robot. This model is not provided by the manufacturer of the robot, and has to be determined experimentally. The transfer function model of the robot and its internal controller were estimated by system identification techniques for a particular configuration of the robot in an earlier study by Aydin et al. \cite{Aydin2020}. The variables $Z_{\mathrm{env}}(s) = F_{\mathrm{env}}(s) / V(s)$ and $Z_h(S) = F_h(s) / V(s)$  in Fig. \ref{fig:control} are the mechanical impedance of the environment and the human operator, respectively (please refer to Aydin et al. \cite{Aydin2020} for the details). The control loop runs at an update rate of 500 Hz.

\subsection{Adaptive control policy}
    \label{sec:adaptivecontrolpolicy}

We divide a pHRI task that can be used for small-batch manufacturing operations such as our drilling task into three well-defined subtasks, Fig. \ref{fig:subtasks}:

\begin{itemize}
    \item{\textbf{Subtask 1, Idle:}} At the beginning, the operator has not grabbed the handle yet, and the system is stationary. This is labeled as Idle.
    \item{\textbf{Subtask 2, Driving:}} The operator is guiding the drill attached to the robot in free space.
    \item{\textbf{Subtask 3, Contact:}} The robot is in contact with the environment and drilling operation is underway.
\end{itemize}

\begin{figure}[b!]
	\includegraphics[width=\columnwidth]{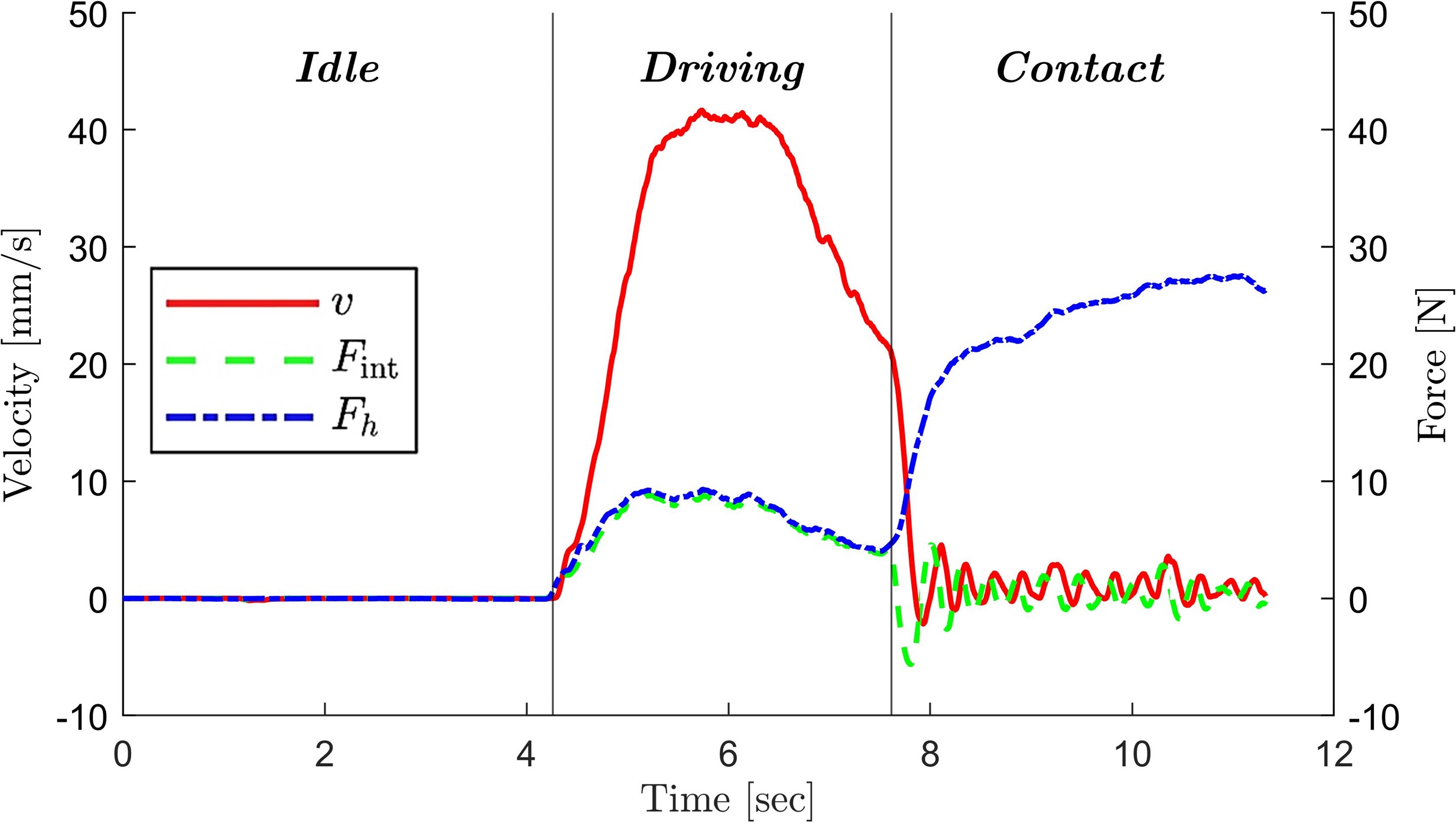}
	\centering
	\caption{Sample time plots of velocity, human and interaction forces for our human-robot collaborative drilling task}
	\label{fig:sampletimeplot}
\end{figure}

Fig. \ref{fig:sampletimeplot} shows the sample time plots of velocity, interaction force, and human force for a single drilling trial, separated into three subtasks. In our implementation, human movement is simply constrained by the robot to the direction perpendicular to the workpiece to simplify the task and conduct a controlled experimental study.

\textit{Idle} corresponds to the few initial seconds at the beginning of every trial when the robot is not moving and is not under the influence of any forces yet. As soon as the operator grabs the handle, \textit{Driving} phase begins. During this subtask, even though the drill itself is active, the robot will only be interacting with the operator, hence human force $F_h$ and interaction force $F_{\mathrm{int}}$ profiles coincide as shown in Fig. \ref{fig:sampletimeplot}. During this subtask, we desire a transparent interaction, so as to minimize human effort during the time the operator is moving the robot towards the workpiece. That is to say, we desire relatively lower values for the damping $b$ in the admittance controller. 

As soon as the drill bit tip touches the workpiece, \textit{Contact} begins, and so does the drilling process. The workpiece starts exerting forces in the opposite direction of the drilling. This is where velocity $v$ and interaction force $F_{\mathrm{int}}$ are relatively low, whereas human force $F_h$ is high (see Fig. \ref{fig:sampletimeplot}). During this phase, the operator pushes against environmental forces to perform the drilling, while the robot is simply moving the drill bit slowly into the workpiece based on the interaction force being fed to the controller. In summary, during the \textit{Contact} phase, human force is greater than the opposing environmental force, and the resulting interaction force ($F_{\mathrm{int}} = F_h - F_{\mathrm{env}}$), which is smaller than both the human and environmental forces, is the input of the admittance controller.

During \textit{Contact}, due to the stiff environment, we desire a stable and robust interaction to make sure safety of the operator is not compromised while the drilling process is underway. For this reason, we desire relatively higher values for the damping $b$ in the admittance controller. Furthermore, our earlier studies \cite{Aydin2018a,Sirintuna2020} showed that by setting the fractional order, $\alpha$, of the admittance controller to a value smaller than 1.0 (see Eq. \ref{eq:eq2}), contact stability can be improved \cite{Aydin2020b}. Hence, in addition to adapting the admittance damping, $b$, throughout the whole task, the fractional order $\alpha$ was also altered during the \textit{Contact} phase.

In order to prevent any sudden or jerky motion at the beginning of \textit{Driving}, we propose to use an intermediate value for damping $b$ in \textit{Idle} phase, a value between the low value chosen for \textit{Driving} and the high value chosen for \textit{Contact}.

\begin{figure}[t]
	\includegraphics[width=\columnwidth]{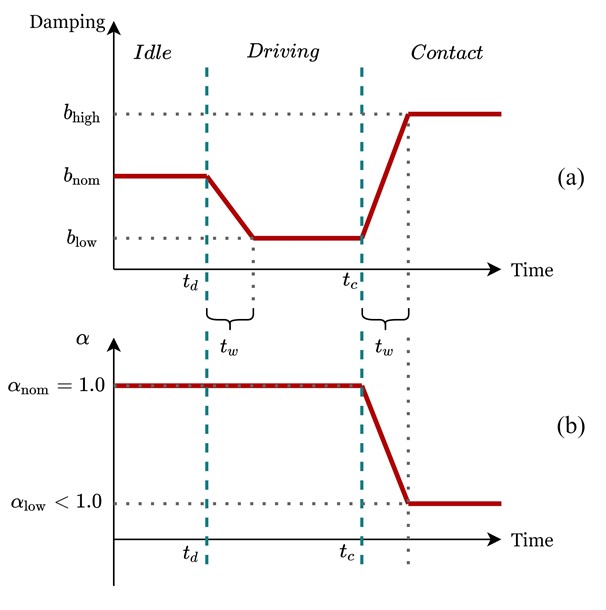}
	\centering
	\caption{The adaptation policy followed in our study for the parameters of the admittance controller; (a) adaptation of damping $b$ in subtask transitions, (b) adaptation of fractional order $\alpha$ at \textit{Contact}.}
	\label{fig:adaptation}
\end{figure}

Adjustment of the controller in subtask transitions should not occur instantaneously, as it would jeopardize the stability of the closed-loop system. Therefore, parameters $b$ and/or $\alpha$ change linearly from the previous value to the new value within a predefined time window $t_w$. We choose a time window of 200 ms for the transition (see Fig. \ref{fig:adaptation}).

In our experimental study, two adaptive controllers were investigated and their performances were compared with that of an admittance controller having fixed parameters. One of them only adjusts damping in order to reduce human effort in the \textit{Driving} phase without compromising stability in the \textit{Contact} phase, while the other one also adapts the $\alpha$ parameter at \textit{Contact} in addition to damping, in order to further improve contact stability during drilling. Further details are given in \hyperref[sec:Experiments]{Section 3}.

\subsection{\revisedP{Subtask processor}}

\begin{figure}[h]
	\includegraphics[width=0.7\columnwidth]{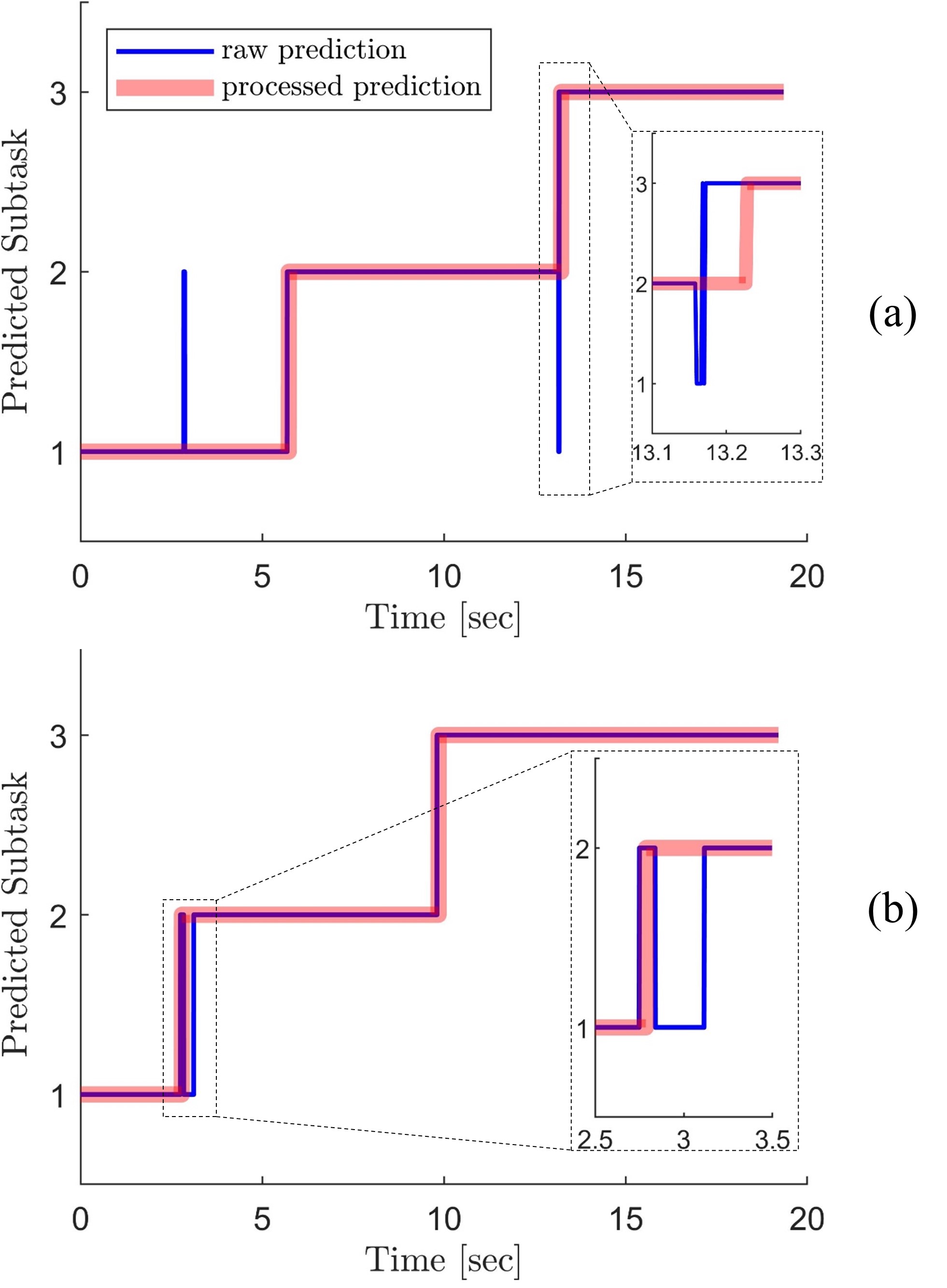}
	\centering
	\caption{Subtask processor in action; (a) The voting buffer eliminates the momentary misclassification of the ANN, (b) The secondary buffer prevents misclassifications in subtask transitions.}
	\label{fig:subtaskprocessor}
\end{figure}

\revisedP{In this paper, we are using a deep learning model to classify subtasks and then adapt control parameters accordingly. Such models are always prone to misclassification in real time applications due to several unpredictable and unavoidable reasons. In order to eliminate any potential instabilities due to misclassification, we add two safety features to our subtask classifier when it is deployed online. These safety features are deployed in the ``Subtask Processor" block of the closed-loop system shown in Fig. \ref{fig:control}. They are detailed below. In addition, Fig. \ref{fig:subtaskprocessor} shows two examples in which the subtask processor improves the classification results.}

\revisedP{
\paragraph{Voting Buffer} This buffer holds the classifier's raw predictions for the previous time steps, and outputs the most frequent prediction in this moving buffer as the \textit{``verdict"} on what the subtask is at the moment. This acts as the equivalent of a low-pass filter for discrete-valued categorical time-series data. Momentary fluctuations in ANN prediction within a subtask can be addressed by this method.
For instance, in Fig. \ref{fig:subtaskprocessor}, there is a momentary misclassification to \textit{Driving} when the drill is turned on, due to a small noise in the data, long before \textit{Driving} actually starts. In addition, at the beginning of \textit{Contact}, there is a momentary misclassification to \textit{Idle}, likely because the subject decelerates to a momentary full halt before commencing drilling, even though the human intention is obviously not \textit{Idle}. The voting buffer prevents the damping value from fluctuating between high and low values, further assuring stability. In our implementation, we have chosen a voting buffer length of 30 time steps, equal to 60 ms in our system.

\paragraph{Secondary Buffer} This buffer stores the verdicts of the voting buffer, and makes sure that whenever the subtask changes from an old value to a new value, it stays in that new value for a given period of time, before it changes again. This is intuitive, as the operator, for instance, will never change their intention immediately after transitioning from one subtask to the other. This safety measure is primarily put in place for the fluctuations during the subtask transitions that are longer than the length of the voting buffer. With this secondary safety feature, once the voting buffer decides to predict \textit{Driving} for the first time for example, the subtask receives a final verdict of \textit{Driving} and stays there for at least 1 second before it can change back into \textit{Idle} again or go to \textit{Contact}. 

The bottom plot in Fig. \ref{fig:subtaskprocessor} shows a case where it takes some time for the operator to fully grasp the handle with their hand. The moment the operator touches the handle, \textit{Driving} has officially begun. However, because of the time that it takes for the operator to fully grab the handle before actually moving the robot, noisy data cause the ANN output to switch to \textit{Driving}, but then return to \textit{Idle} for some time, before going back to \textit{Driving} again as soon as the motion starts. Note that this fluctuation is happening in a longer period of time than the voting buffer length. This is where the secondary buffer comes in handy, and prevents returning back to \textit{Idle} after \textit{Driving} has already been detected by the voting buffer in the previous time step. The voting buffer is for momentary fluctuations that can happen anywhere, while the secondary buffer is for confusions that can occur only in subtask transitions, and for longer periods of time.

We must also keep in mind that we cannot simply increase the length of the voting buffer, as this would create long delays between actual and predicted subtask transitions. The secondary buffer induces no delay in subtask classifier's operation; it simply makes sure that once the subtask transitions to a new value, it will not change for some period of time.

\vspace{10pt}

The output of the secondary buffer is accepted as the final verdict of the predicted subtask, denoted as ``Processed Label" in Fig. \ref{fig:control}, which is fed to the adaptation mechanism. The adaptation mechanism then sets the control parameters (admittance damping and the fractional order) accordingly.}

\subsection{ANN model}

Detecting the subtask is considered as a time-series classification problem in our context. Due to highly nonlinear relationships that exist between the inputs and the outputs in pHRI scenarios such as that of ours, deep learning methods are chosen as the model family, as they have proven to be highly effective for such purposes. One very popular and capable model for time-series classification is the Artificial Neural Network (ANN). ANN models are capable of learning all manner of highly nonlinear relationships in dynamic systems \cite{IanGoodfellowYoshuaBengio2016}. We therefore developed a fully-connected feed-forward multi-layer artificial neural network model for subtask classification. As typical in time-series problems, our inputs are sequences of data collected from previous time steps, and our output is the predicted subtask at the current time step.

We choose Softmax to be the activation function of our output layer, as it is typical in classification problems, and ReLU (Rectified Linear Unit) as the activation function of the hidden layers, as it is one of the most popular activation functions for neural networks \cite{IanGoodfellowYoshuaBengio2016}.

Since our inputs are in fact sequences of data, each time step in our sequence is considered as a separate input. Consequently, the input size of the ANN depends on the length of the input sequence, which itself is another hyperparameter for us. That is to say, input size to the ANN is the number of features multiplied by the sequence length.

We select ADAM \cite{Kingma2015} as the optimization algorithm for training the model, validation accuracy as the main metric, and cross-entropy as the cost function to be minimized. For preventing over-fitting, to which deep learning models are very vulnerable, we include a dropout layer \cite{Srivastava2014} after each hidden layer, as well as applying $L_2$ regularization. A maximum number of 80 epochs and a minibatch size of 128 are chosen during training, along with a learning rate of 0.001, after initialization with the Glorot-uniform method. The Keras deep learning library is used in Python for training.

After a close inspection of drilling experiments, we realize that the most prominent features in determining and distinguishing subtasks from one another are velocity $v$, interaction force $F_{\mathrm{int}}$, and human force $F_h$, as shown in Fig. \ref{fig:sampletimeplot}. Therefore, we choose these three features for training our ANN model. In our initial attempts, only the velocity and interaction force were selected as the features for the model, but they often produced sub-par results in comparison to the three features chosen for the current implementation. \hyperlink{section.5}{Section 5} provides further detail on this matter.

For labeling the data, \textit{Idle} starts at the beginning of the task when the velocity is zero. We label the beginning of \textit{Driving} as the first time human force, $F_h$, exceeds some threshold value. \textit{Contact} starts when the drill bit touches the workpiece and drilling begins. Human and environment forces part ways from each other, and there is often a distinct sudden deceleration in movement at the beginning of \textit{Contact}. The \textit{Contact} is therefore labeled visually by inspecting the data. The experiment ends when the target drill depth is reached.

\begin{table}[t]
\caption{Hyperparameters chosen for the ANN model}
    \vspace{3pt}
	\centering
	\resizebox{\columnwidth}{!}{%
	\begin{tabular}{ll}
	\textbf{Hyperparameter}                         & \textbf{Chosen Value} \\ \hline
	Number of hidden layers                         & 5                     \\ 
	Number of units on every hidden layer           & 75                     \\ 
	Sequence length (number of time steps in input) & 60                     \\  
	Dropout rate                                    & 0.1                     \\
	L\textsubscript{2} regularization parameter     & 0.001                    
	\end{tabular}%
	}
	\label{tab:hyperparams}
\end{table}

During the training stage of the model, 75\% of the collected data was used as the training set, and the remaining 25\% was used for validation. Different ANN configurations were trained, and the configuration shown on Table \ref{tab:hyperparams} was chosen eventually.

\begin{table*}[t!]
\caption{Overview of the training and testing experiments. Note that mass $m$ is expressed in $\mathrm{Kg.s^{\alpha - 1}}$ and damping $b$ is expressed in Ns/m.}
\vspace{3pt}
\centering
\resizebox{0.85\textwidth}{!}{%
\begin{tabular}{|lll|}
\hline
\textbf{Experiment}                                                    & \textbf{Training} & \textbf{Testing} \\ \hline
\textbf{Description} &
  \begin{tabular}[c]{@{}l@{}}Performed by Sirintuna et al. \cite{Sirintuna2020}\\ using position-based adaptive control\end{tabular} &
  \begin{tabular}[c]{@{}l@{}}Adaptive control using \\ subtask classification with ANN\end{tabular} \\ \hline
\textbf{No. Subjects}                                                 & 7                                & 12                \\ \hline
\multirow{3}{*}{\textbf{\begin{tabular}[c]{@{}l@{}}Admittance\\ Controllers\end{tabular}}} &
  Fixed IOAC: $m = 30$, $b = 2250$, $\alpha=1.0$ &
  C1 - Fixed IOAC: $m=50$, $b=400$, $\alpha=1.0$ \\ \cline{2-3} 
 &
  \begin{tabular}[c]{@{}l@{}}Adaptive IOAC: $m=69$, $b=711$ at the\\ beginning,  continuously changes with position\\ toward $m = 30$, $b=2250$ at contact.\end{tabular} &
  \begin{tabular}[c]{@{}l@{}}C2 - Adaptive IOAC: $m=50$, $\alpha = 1.0$,\\ $b$ changes in subtask transitions. \end{tabular} \\ \cline{2-3} 
 &
  \begin{tabular}[c]{@{}l@{}}Adaptive FOAC: $m=69$, $b=711$,\\ $\alpha$ starts at 1.0 and changes continuously \\ toward 0.85 at contact.\end{tabular} &
  \begin{tabular}[c]{@{}l@{}}C3 - Adaptive FOAC: $m=50$, \\ $b$ and $\alpha$ change in subtask transitions.\end{tabular} \\ \hline
\textbf{\begin{tabular}[c]{@{}l@{}}Total No.\\ Trials\end{tabular}}   & 84                               & 144              \\ \hline
\textbf{$m$ range}                                                    & 30-69                            & 50               \\ \hline
\textbf{$b$ range}                                                    & 711-2250                         & 200-400          \\ \hline
\textbf{$\alpha$ range}                                               & 0.85-1.0                         & 0.85-1.0         \\ \hline
\textbf{\begin{tabular}[c]{@{}l@{}}Workpiece\\ Material\end{tabular}} & Plywood                          & Cardboard        \\ \hline
\textbf{\begin{tabular}[c]{@{}l@{}}Drillbit\\ RPM\end{tabular}}       & $\approx 2000$                   & $\approx 4500$   \\ \hline
\end{tabular}%
}
\label{tab:experiments}
\end{table*}

\section{Experiments}
	\label{sec:Experiments}

The goal of our experimental study is to demonstrate the benefits of proposed approach in a realistic pHRI scenario. Collaborative drilling is selected as the representative pHRI task in our study. We implement an adaptive admittance controller to minimize human effort during the \textit{Driving} phase while improving stability in the \textit{Contact} phase based on the subtask classification. To this end, firstly, a training dataset was required for the deep learning model. After the model was trained, another set of testing experiments was required for validating the efficacy of the trained model under different circumstances. \revisedSecondRevision{Table~\ref{tab:experiments}} summarizes the experiments along with their details.

\begin{itemize}
    \item \textbf{Training experiments:} The experiments were performed by Sirintuna et al. \cite{Sirintuna2020} using an admittance controller with no adaptation or manual adaptation; 7 subjects, 3 control parameter settings, 4 trials each
    
    \item \textbf{Testing experiments:} New experiments were performed in this study using an adaptive admittance controller with ANN subtask classifier; 12 new subjects, 3 control parameter settings, 4 trials each
\end{itemize}

\subsection{Training Experiments}
The data acquired by Sirintuna et al. \cite{Sirintuna2020} was used as the training set for our study. The experiments conducted in the aforementioned study were for a drilling task similar to ours, performed with the same collaborative robot.  In the referenced study, three different admittance controllers, as detailed below, were used, each tried with 7 participants (6 male and 1 female with an average age of 24.9 years) performing 4 trials per controller. Hence, each participant performed 12 trials (3 controllers x 4 repetitions) and there were a total of 84 trials (7 participants x 12 trials) in the training experiments. The order of the trials was randomized while the same order was displayed to each participant. 

\paragraph{Protocol} The participant waited for the light on the screen of an AR goggle to turn green. They then grabbed the handle, moved the robot towards the workpiece at a natural pace, and drilled a hole on the workpiece. The AR goggle showed the distance to the workpiece, as well as penetration depth, to the participant. The participant was instructed to drill a depth of 5 mm into the workpiece.

\paragraph{Admittance controllers} Three admittance controllers were used in the training experiments:
\begin{enumerate}[a)]
	\item \textit{Fixed IOAC} with $m = 30\,\mathrm{Kg}$,\\$b=2250\,\mathrm{Ns/m}$
	\item \textit{Adaptive FOAC} with $m = 69\,\mathrm{Kgs^{\alpha - 1}}$, $b = 711\,\mathrm{Ns/m}$; $\alpha$ started at 1.0 and decreased linearly with position, to 0.85 at contact.
	\item \textit{Adaptive IOAC} where $m = 69\,\mathrm{Kg}$, $b = 711\,\mathrm{Ns/m}$ were the values at the beginning, and they were altered continuously according to the impedance matching with the adaptive FOAC above, to $m = 30\,\mathrm{Kg}$, $b = 2250\,\mathrm{Ns/m}$ at contact.
\end{enumerate}

\subsection{Testing Experiments}

Further experimentation was required to evaluate the performance of the trained model. Model performance in these experiments determined if the model was sufficiently robust and flexible under different conditions. Hence, we selected different participants, admittance controllers, adaptation policies, and a workpiece with different material properties for the testing experiments from the training experiments to investigate the performance of the subtask classifier.

We performed the testing experiments with 12 participants (10 male and 2 female with an average age of 24.67$\pm$2.05 years). Three different admittance controller conditions (C1, C2, and C3) were tested as detailed below. Each condition was repeated 4 times. Hence, each participant performed 12 trials (3 controllers x 4 repetitions) and there were a total of 144 trials (12 participants x 12 trials) in the testing experiments. The order of the trials was randomized while the same order was displayed to each participant.

\paragraph{Protocol} The protocol followed in the testing experiments was similar to that of the training experiments. Instead of an AR goggle as in the training experiments, a computer monitor was used for providing visual feedback and guidance to the participant. The participant was asked to stand inside a marked space, and waited for the word ``START" to appear on the computer monitor in front of them. At this point the participant grabbed the handle, and moved it towards the workpiece, (\textit{Driving}). The word ``START" disappeared from the screen at this moment. Once the robot contacted the workpiece and drilling started (\textit{Contact}), a progress bar appeared on the monitor, displaying the drill depth in real-time. The target drill depth was chosen as 5 mm as in the training experiments. Once the target drill depth was achieved, the word ``RETRACT" appeared on the monitor, at which point the participant pulled the drill out of the workpiece. The detection of \textit{Driving} and \textit{Contact}, which triggered a change in the visual information displayed on the monitor, was based on the processed label, outputted by the subtask processor, which in turn was tied to the ANN model as shown in \revisedSecondRevision{Fig.~\ref{fig:control}.} 
\vspace{5pt}

\paragraph{Admittance controllers}
We conducted the testing experiments under three different admittance controllers. For establishing a baseline for comparison, we used an IOAC with fixed admittance damping. In addition, we designed an adaptive IOAC utilizing the adaptation policy explained in \hyperref[sec:adaptivecontrolpolicy]{Section 2.3} for the admittance damping, $b$ as shown in Fig. \ref{fig:adaptation}a. Finally, for investigating the additional benefits of adjusting the fractional order, especially for the \textit{Contact} phase, we also designed an adaptive FOAC utilizing the adaptation policy explained in \hyperref[sec:Approach]{Section 2} for the admittance damping $b$ as shown in Fig. \ref{fig:adaptation}a and the fractional order, $\alpha$, as shown in Fig. \ref{fig:adaptation}b. The parameter values chosen for the controllers are tabulated in Table \ref{tab:dampvals}. The details for those selections are explained below. 

\begin{table}[t]
    \caption{Parameters of the admittance controllers used in the testing experiments. Note that only the admittance damping, $b$, was adapted in C2 while both the admittance damping, $b$, and fractional order, $\alpha$, were adapted in C3.}
    \vspace{3pt}
	\centering
	\resizebox{\columnwidth}{!}{
	\begin{tabular}{lll}
	\hline
	\textbf{Parameter} & \textbf{Chosen Value}     & \textbf{Unit} \\ \hline
	\multicolumn{3}{l}{C1 - Fixed IOAC}                                 \\ \hline
	$m$                & 50                        & Kg            \\ 
	$b$                & 400                       & Ns/m          \\ 
	$\alpha$                & 1.00                       &           \\\hline
	\multicolumn{3}{l}{C2 - Adaptive IOAC: $b$ was altered at subtask transitions}                              \\ \hline
	$m$                & 50                        & Kg            \\ 
	$b_{\mathrm{low}}$          & 200                       & Ns/m          \\ 
	$b_{\mathrm{nom}}$          & 300                       & Ns/m          \\ 
	$b_{\mathrm{high}}$         & 400                       & Ns/m          \\ 
	$t_w$              & 200                       & ms            \\ 
	$\alpha$              & 1.00                       &             \\\hline
	\multicolumn{3}{l}{C3 - Adaptive FOAC: $b$ and $\alpha$ were altered at subtask transitions}                              \\ \hline
	$m$, $b$, $t_w$           & Same as C2 &               \\ 
	$\alpha_{\mathrm{nom}}$           & 1.00                      &               \\
	$\alpha_{\mathrm{low}}$           & 0.85                      &               \\\hline
	\end{tabular}}
	\label{tab:dampvals}
\end{table}

\begin{figure}[t]
	\includegraphics[width=0.8\columnwidth]{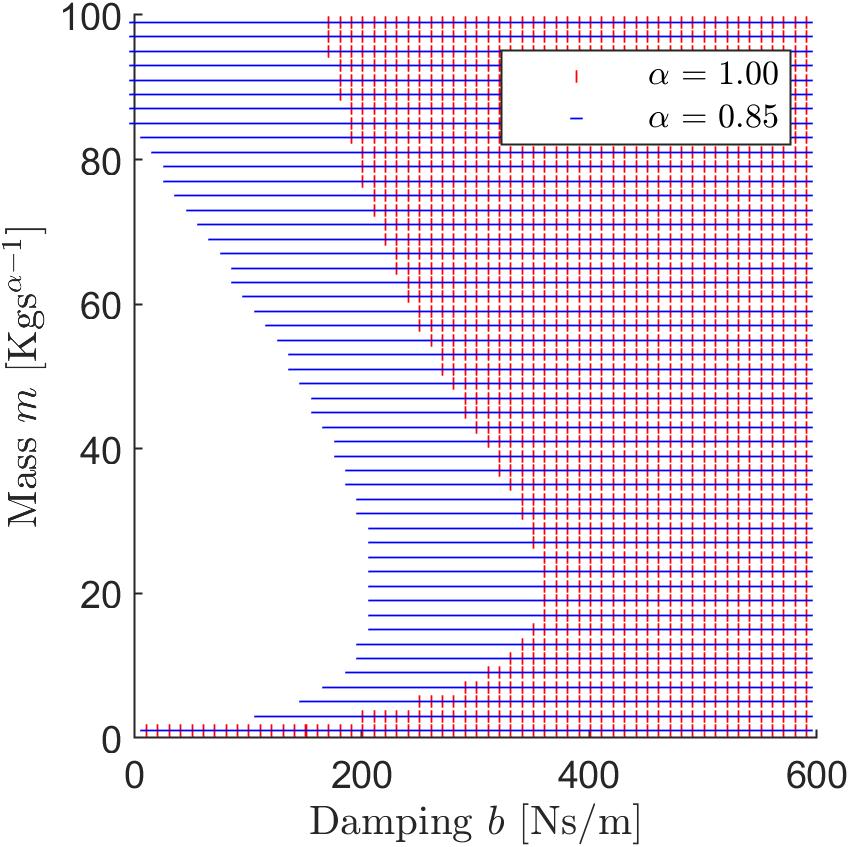}
	\centering
	\caption{Stability map for the \textit{Contact} phase under IOAC ($\alpha=1.00$) and FOAC  ($\alpha=0.85$). \revisedS{Note that the stable regions are the colored regions.}}
	\label{fig:stabilitymap}
\end{figure}

In order to choose appropriate control parameters, knowledge of stability thresholds for the closed-loop control system was required, which were extracted from its stability map. A stability map is a graphical representation of the
controller parameters for which the resulting closed-loop system becomes stable for a range of environment/human arm impedance values (see the details in \cite{Aydin2018a}). The approach followed by Aydin et al.~\cite{Aydin2020} was utilized to conduct the stability analysis and obtain the stability map for the closed-loop system shown in Fig. \ref{fig:control}. A fairly conservative stability map for the \textit{Contact} phase, which is more susceptible to instability, is shown in Fig. \ref{fig:stabilitymap} for our pHRI system. According to this map, in the \textit{Contact} phase, with an admittance mass value of $m = 50\, \mathrm{Kg}$, the stability threshold for the damping value under IOAC was somewhere between 250 and 300 Ns/m. In order to account for noise as well as unpredictable human or environmental dynamics during drilling, a more conservative damping value of 400 [Ns/m] was selected for \textit{Contact}. Obviously, this damping value was high for the \textit{Driving} phase where the robot was simply being guided by the human in free space, and would increase human effort unnecessarily. This is why an adaptive control has been suggested in this study in the first place. For the \textit{Driving} phase, a reasonable and conservative damping value of 200 Ns/m was chosen. As mentioned earlier, in order to avoid any jerky motion when the user grabbed the handle, a medium damping value of 300 Ns/m was selected for the \textit{Idle} phase, a value that fell in between the damping values selected for the \textit{Driving} and \textit{Contact} phases.

In summary, the following admittance controllers were used in the testing experiments:
\begin{enumerate}[a)]
	\item \textit{C1 - Fixed IOAC:} $m = 50\,\mathrm{Kg}$, $b=400\,\mathrm{Ns/m}$, $\alpha = 1.00$
	\item \textit{C2 - Adaptive IOAC:} $m = 50\,\mathrm{Kg}$, $b \in [200,\,400]\,\mathrm{Ns/m}$, $\alpha=1.00$
	\item \textit{C3 - Adaptive FOAC:} $m = 50\,\mathrm{Kg.s^{\alpha - 1}}$, $b \in [200,\,400]\,\mathrm{Ns/m}$, $\alpha \in [0.85,1.00]$
\end{enumerate}

The values of controller parameters used in the testing experiments are tabulated in Table \ref{tab:dampvals} in detail, and can be compared against the parameter values used in the training experiments in Table \ref{tab:experiments}.

\begin{figure*}[h]
	\includegraphics[width=\textwidth]{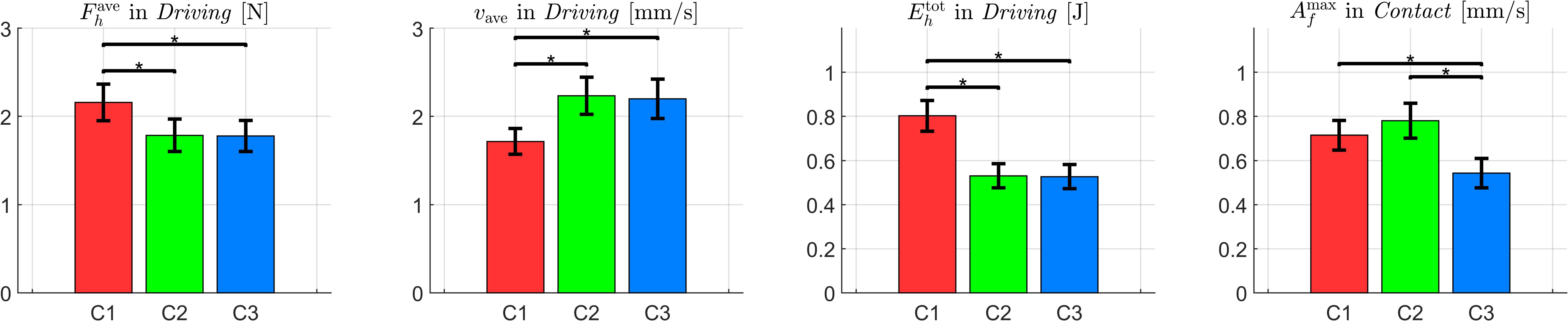}
	\centering
	\caption{Comparison of performance metrics for the testing experiments; (C1) fixed IOAC, (C2) adaptive IOAC, (C3) adaptive FOAC; note that the bars represent the mean values of the normalized metrics for all trials of all participants, and the error lines represent the 95\% confidence intervals. Horizontal lines with asterisks denote statistically significant pairwise comparisons with $p = 0.001$ as the threshold.}
	\label{fig:perfmetrics}
\end{figure*}

\begin{figure}[h]
	\includegraphics[width=\columnwidth]{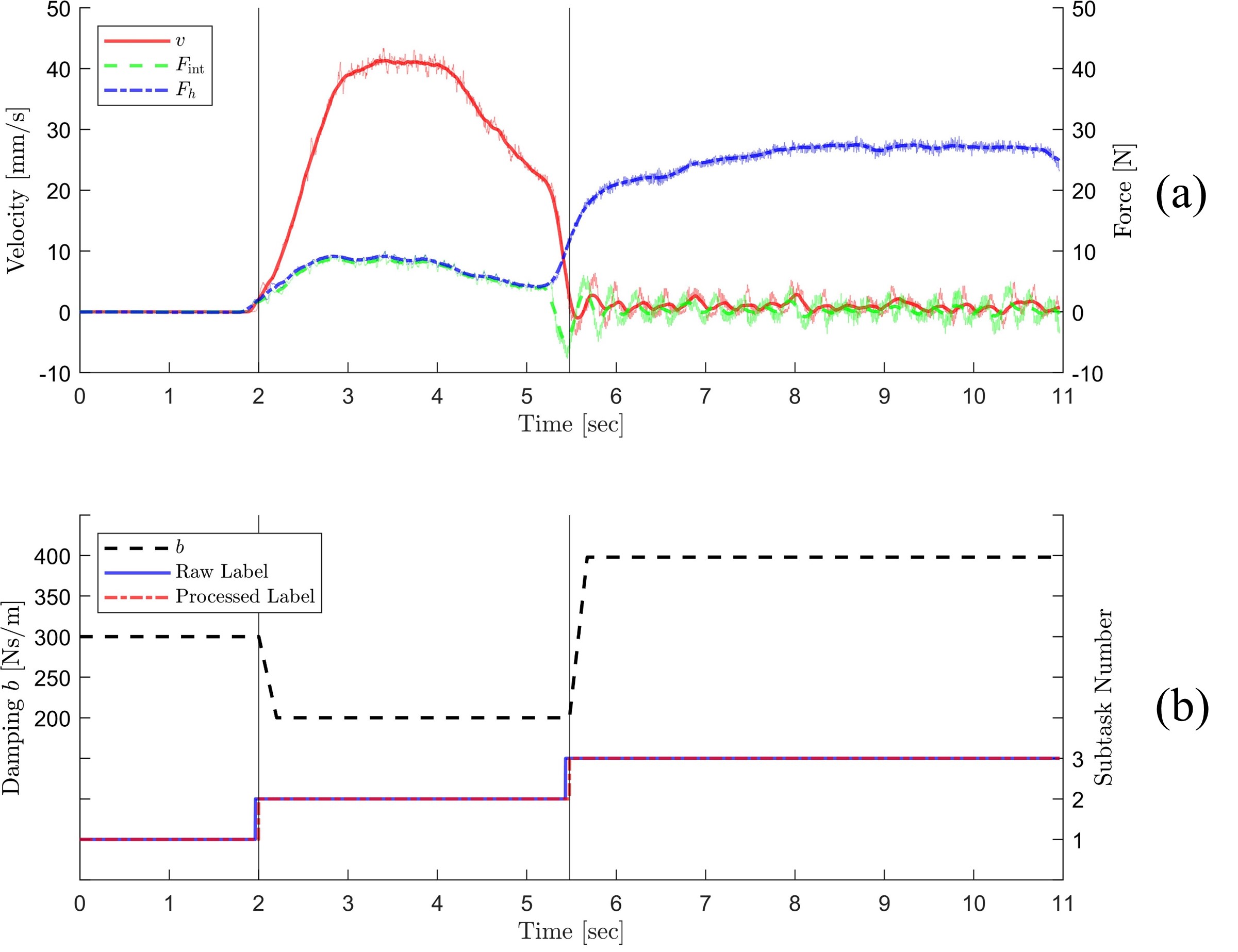}
	\centering
	\caption{(a) Sample time plots of velocity, human and interaction force recorded under adaptive IOAC (C2); (b) the raw (solid blue) and processed (dot-dashed red) outputs of the classifier for this trial. The dashed black line shows the adaptation in admittance damping. Note that the subtasks have been separated by vertical lines in both plots.}
	\label{fig:sampleseries2}
\end{figure}

\section{Results}
	\label{sec:Results}

In this section, the results of the subtask classifier and the performances of the three admittance controllers are reported.

\subsection{Subtask classifier performance}

Table \ref{tab:confmat_all} reports the performance of the classifier. Accuracy levels, weighted $F_1$ scores, and normalized confusion matrices are reported in this table. Note that all values are expressed in percentages. Sample time plots of velocity, human and interaction forces recorded under adaptive IOAC (C2) is shown in Fig. \ref{fig:sampleseries2}, which also displays the raw and processed labels of the classifier. The adaptation of admittance damping is also depicted in the same figure.

\begin{table}[t]
\caption{Performance of the subtask classifier under all admittance controllers. All values are expressed in percentage.}
\vspace{3pt}
\centering
\resizebox{\columnwidth}{!}{%
\begin{tabular}{ccccccc}
\textbf{Controller} &
  \textbf{Accuracy} &
  \textbf{$F_1$ Score} &
  \multicolumn{4}{c}{\textbf{Confusion Matrix}} \\ \hline
\multirow{5}{*}{C1} &
  \multirow{5}{*}{97.87} &
  \multirow{5}{*}{97.86} &
  \multirow{2}{*}{\begin{tabular}[c]{@{}c@{}}Actual\\ Subtask\end{tabular}} &
  \multicolumn{3}{c}{Predicted Subtask} \\
 &
   &
   &
   &
  1 &
  2 &
  3 \\
 &
   &
   &
  1 &
  100 &
  0 &
  0 \\
 &
   &
   &
  2 &
  1.48 &
  98.52 &
  0 \\
 &
   &
   &
  3 &
  0 &
  4.55 &
  95.45 \\ \hline
\multirow{5}{*}{C2} &
  \multirow{5}{*}{97.15} &
  \multirow{5}{*}{97.12} &
  \multirow{2}{*}{\begin{tabular}[c]{@{}c@{}}Actual\\ Subtask\end{tabular}} &
  \multicolumn{3}{c}{Predicted Subtask} \\
 &
   &
   &
   &
  1 &
  2 &
  3 \\
 &
   &
   &
  1 &
  100 &
  0 &
  0 \\
 &
   &
   &
  2 &
  2.84 &
  97.16 &
  0 \\
 &
   &
   &
  3 &
  0 &
  4.90 &
  95.10 \\ \hline
\multirow{5}{*}{C3} &
  \multirow{5}{*}{97.81} &
  \multirow{5}{*}{97.78} &
  \multirow{2}{*}{\begin{tabular}[c]{@{}c@{}}Actual\\ Subtask\end{tabular}} &
  \multicolumn{3}{c}{Predicted Subtask} \\
 &
   &
   &
   &
  1 &
  2 &
  3 \\
 &
   &
   &
  1 &
  100 &
  0 &
  0 \\
 &
   &
   &
  2 &
  1.68 &
  98.32 &
  0 \\
 &
   &
   &
  3 &
  0 &
  4.37 &
  95.63
\end{tabular}%
}
\label{tab:confmat_all}
\end{table}

According to Table \ref{tab:confmat_all}, the subtask classifier performs well on the testing data despite major differences between the training and testing experiments in terms of setup, environment (i.e. the material properties of the workpiece), participants and admittance controllers. This result further justifies the application of a learning-based subtask classification for adaptive control over a rule-based one. As it can be observed from this table, the classifier accuracy is around 98\% under all controllers while the largest confusion is between \textit{Driving} and \textit{Contact}.

The delay in subtask detection was approximately 235 ms on average, which is acceptable since it is rather short compared to the amount of time it would take for the potential instabilities to build up at the \textit{Contact} phase under some low damping. This issue is further discussed in \hyperref[sec:Discussion]{Section 5}.

\subsection{Controller performance}

We compared the participants' performances for the controllers tabulated in Table \ref{tab:dampvals}. The following performance metrics were considered for this comparison:

\begin{itemize}
    \item Average human force in the \textit{Driving} phase,\\ $F_h^{\mathrm{ave}}=\frac{1}{t_c-t_d}\int_{t_d}^{t_c}F_h(t)\,dt,\:\mathrm{[N]}$
    \item Average velocity in the \textit{Driving} phase,\\ $v_{\mathrm{ave}} = \frac{1}{t_c-t_d}\int_{t_d}^{t_c}v(t)\,dt,\:\mathrm{[m/s]}$
    \item Total Human Effort in the \textit{Driving} phase,\\ \hl{$E_h^{\mathrm{tot}} = \int_{t_d}^{t_c} |F_h(t)\,v(t)|\,dt,\:\mathrm{[J]}$}
    \item Peak oscillation amplitude of the end-effector velocity in the \textit{Contact} phase, $A_f^{\mathrm{max}}$
\end{itemize}

In the expressions above, $t_d$ and $t_c$ are the starting times for \textit{Driving} and \textit{Contact} respectively, and $A_f$ is the single-sided amplitude spectrum as a function of frequency $f$, calculated by the Fourier transform.
Human force, velocity and especially total human effort in \textit{Driving} phase provide us with valuable information as to how transparent the robot is to the user during free motion, and peak oscillation amplitude of velocity during drilling provides useful information as to how much the stability robustness is compromised under different controller conditions.

\revisedS{The calculation of $A_f^\mathrm{max}$ was performed using the Cartesian velocity of the end-effector for \textit{Subtask 3: Contact}. For each trial, the data corresponding to the \textit{Contact} phase was extracted, denoised and detrended by a band-pass filter with a frequency range of 1.0 to 20.0 Hz (note that the drill operating frequency is $\approx$75 Hz), then output of its FFT was converted to the single-sided amplitude spectrum, from which the peak oscillation amplitude ($A_f^\mathrm{max}$) was obtained.}

The mean values of performance metrics were calculated for each participant under each controller condition (C1, C2, C3). The population mean and its 95\% confidence interval are reported in Fig \ref{fig:perfmetrics}. \revisedP{In order to investigate the effect of control conditions on performance metrics, one-way ANOVA (analysis of variance) with repeated measures was performed, while considering the type of admittance controller as the main factor, with $p = 0.001$ for testing the null hypothesis. Fig \ref{fig:perfmetrics} reports the results of this analysis.}

\begin{figure}[t]
	\includegraphics[width=\columnwidth]{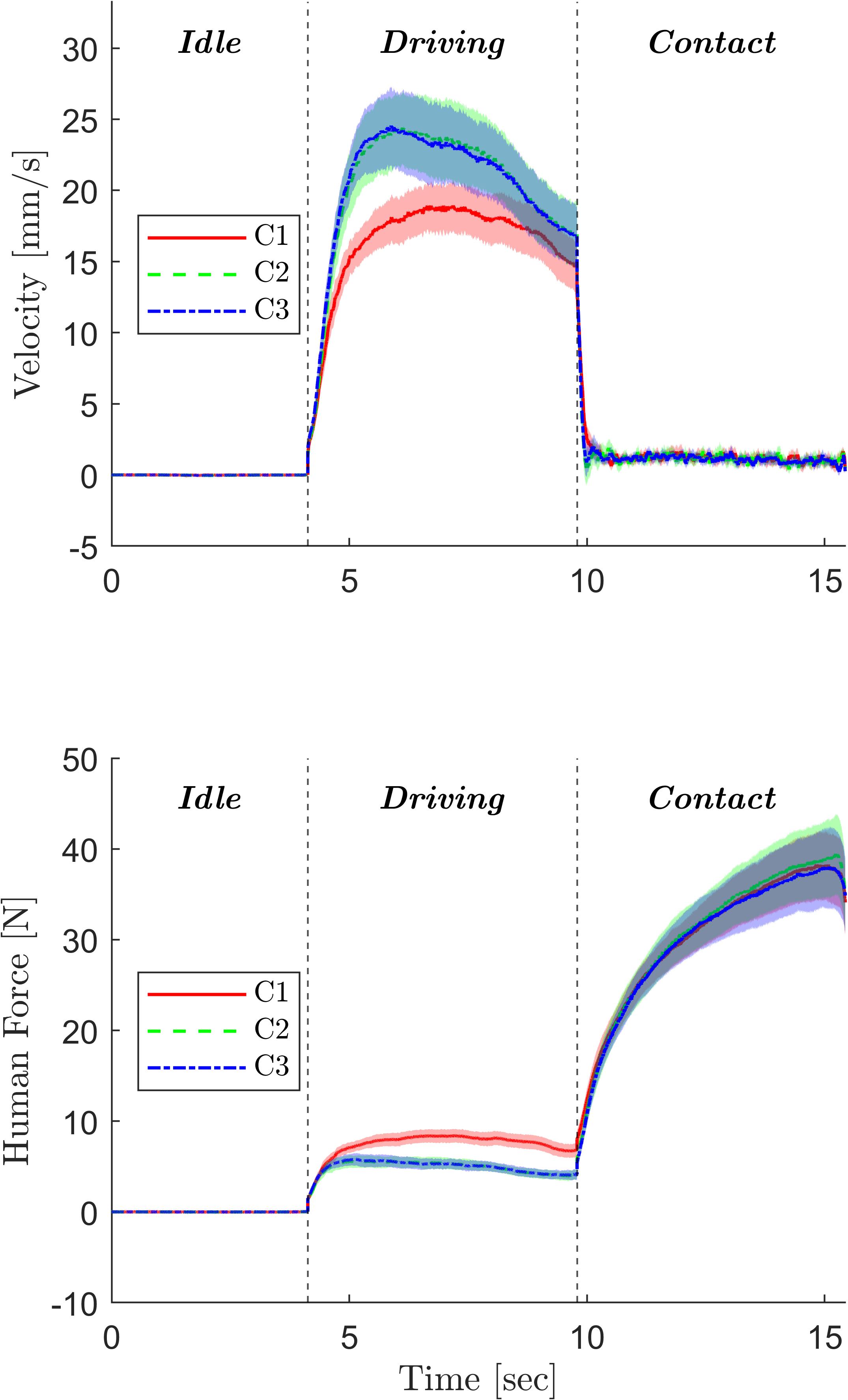}
	\centering
	\caption{Time plots of velocity and human force under all admittance controllers. The dark colored curves are the mean values of all participants, and the shaded regions surrounding them are the 95\% confidence intervals.}
	\label{fig:shadedtimeplots}
\end{figure}

Fig. \ref{fig:perfmetrics} reports the performance metrics for fixed IOAC (C1) compared to those for the adaptive controllers (C2, C3). The results show lower human force, lower human effort and higher velocity in the \textit{Driving} phase for the adaptive controllers compared to the fixed controller. This is because the damping $b$ was lower in the \textit{Driving} phase under the adaptive controllers. The robot was more responsive to input, hence more transparent. The participants could move the robot easily at higher velocities and with less effort. Higher velocities also mean lower task completion times, leading to higher task efficiency in the long run. These improvements with adaptive control in comparison to fixed control prove that adaptation based on subtask classification increases overall task efficiency in pHRI tasks that involve contact with stiff environments.

Note that the damping value chosen for \textit{Driving}, $b_{\mathrm{low}}$, was not applicable for the whole task, because it was lower than the stability thresholds of the \textit{Contact} phase, according to Fig. \ref{fig:stabilitymap}. On the other hand, as shown in Fig. \ref{fig:perfmetrics}, a fixed damping of 400 Ns/m for the whole task (C1) increased human effort in \textit{Driving}, and hence decreased the task efficiency. Consequently, it is shown that an adaptive interaction controller with subtask classification is more efficient than an interaction controller with fixed parameters. Moreover, Fig. \ref{fig:perfmetrics} shows that peak oscillation amplitudes in \textit{Contact} was lower for C3 than those for C1 and C2, indicating that the system was more stable when fractional order $\alpha$ was less than 1.0. A closer look at the stability maps in Fig. \ref{fig:stabilitymap} confirms this observation, as stable regions for lower $\alpha$ value (0.85) are much larger than those for IOAC ($\alpha = 1.0$). Furthermore, this result provides evidence that FOAC offers a better trade-off between stability and transparency, allowing us to choose wider ranges of admittance damping values without compromising stability. This result is inline with the results reported by Sirintuna et al. \cite{Sirintuna2020} and Aydin et al. \cite{Aydin2020,Aydin2020b}. \revisedS{It must be noted that even though it is possible to achieve similar oscillation metrics during \textit{Contact} using an IOAC with a smaller mass/damping ratio, such a provision would require a relatively higher damping value to make the ratio smaller, which would decrease transparency during \textit{Driving}, resulting in higher effort and slower motion. One of the main advantages of FOAC over IOAC is that it maintains stability \textit{without} sacrificing transparency (see \cite{Aydin2018a} for a detailed comparison of IOAC and FOAC). }  

Fig. \ref{fig:shadedtimeplots} demonstrates the comparative time plots of different control schemes used in the testing experiments. For eliminating the differences in the time duration among different trials, the data was time-normalized and interpolated. This resulted in the displayed data for every subtask to have the same length among all trials, after which statistical data such as means and standard deviations could fairly be extracted. The differences in velocity and human force magnitudes under the three admittance controllers can be observed in this figure.

\section{Discussion}
	\label{sec:Discussion}
In this study, we showed that a pHRI task can be divided into multiple subtasks and when these subtasks are detected by deep learning techniques in real-time, appropriate control parameters can be assigned to each subtask to improve the task efficiency and contact stability. The results in Fig. \ref{fig:perfmetrics} showed that the total human effort in \textit{Driving} was 20$\%$ lower under the adaptive controllers (C2 and C3) compared to the fixed controller (C1), and when the fractional order $\alpha$ was also adapted (C3) at \textit{Contact}, peak oscillation amplitude in \textit{Contact} was 25$\%$ smaller compared to the fixed admittance controller (C1).

\begin{figure}[t]
	\includegraphics[width=\columnwidth]{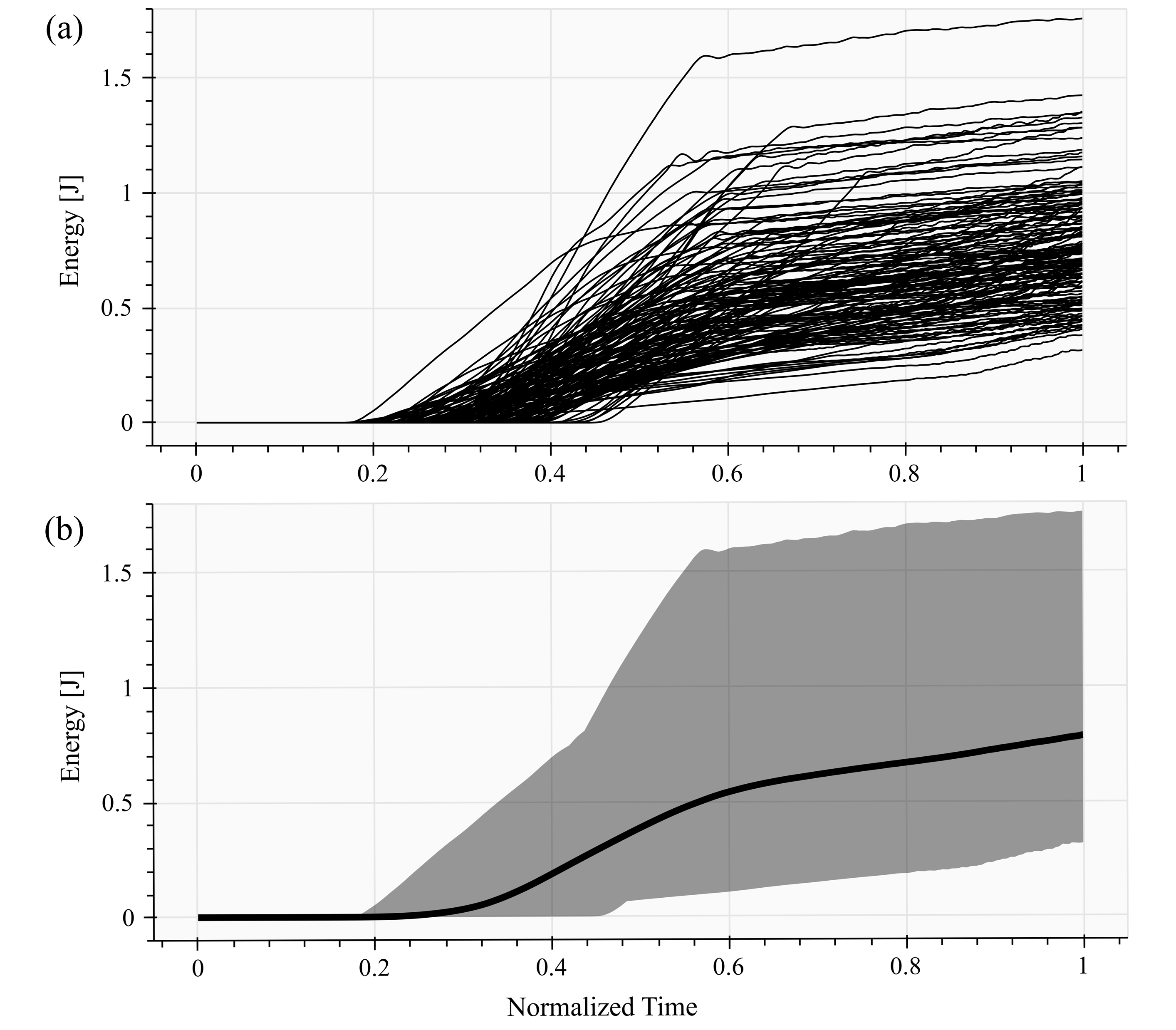}
	\centering
	\caption{a) Exchanged energy in all trials of all subjects; b) The range of exchanged energy for all trials (shaded area) and its mean (solid curve). Data is normalized across time.}
	\label{fig:passivity}
\end{figure}


\revisedSecondRevision{
Altering the control parameters in real time may jeopardize the stability of a pHRI system. We monitored the energy exchange in all trials of our subjects to see if passivity~\cite{Hannaford2002,Hwan2004} is violated in any of them, as shown in Fig.~\ref{fig:passivity}. This figure suggests that we did not run into any case where the passivity was violated during the execution of the experiments, indicating that the system dissipated energy rather than generating it, thereby showing that no instability was observed during the experiments. However, this is not sufficient to claim stability for all possible conditions in general. One might implement stability/passivity observers~\cite{Hannaford2002,Hwan2004,CAMPEAULECOURS201645,Okunev2012,Dimeas2016} to track energy exchange within the system or the amplitude of oscillations and then make corrective actions accordingly for maintaining the stability.

\begin{figure}[h]
	\includegraphics[width=\columnwidth]{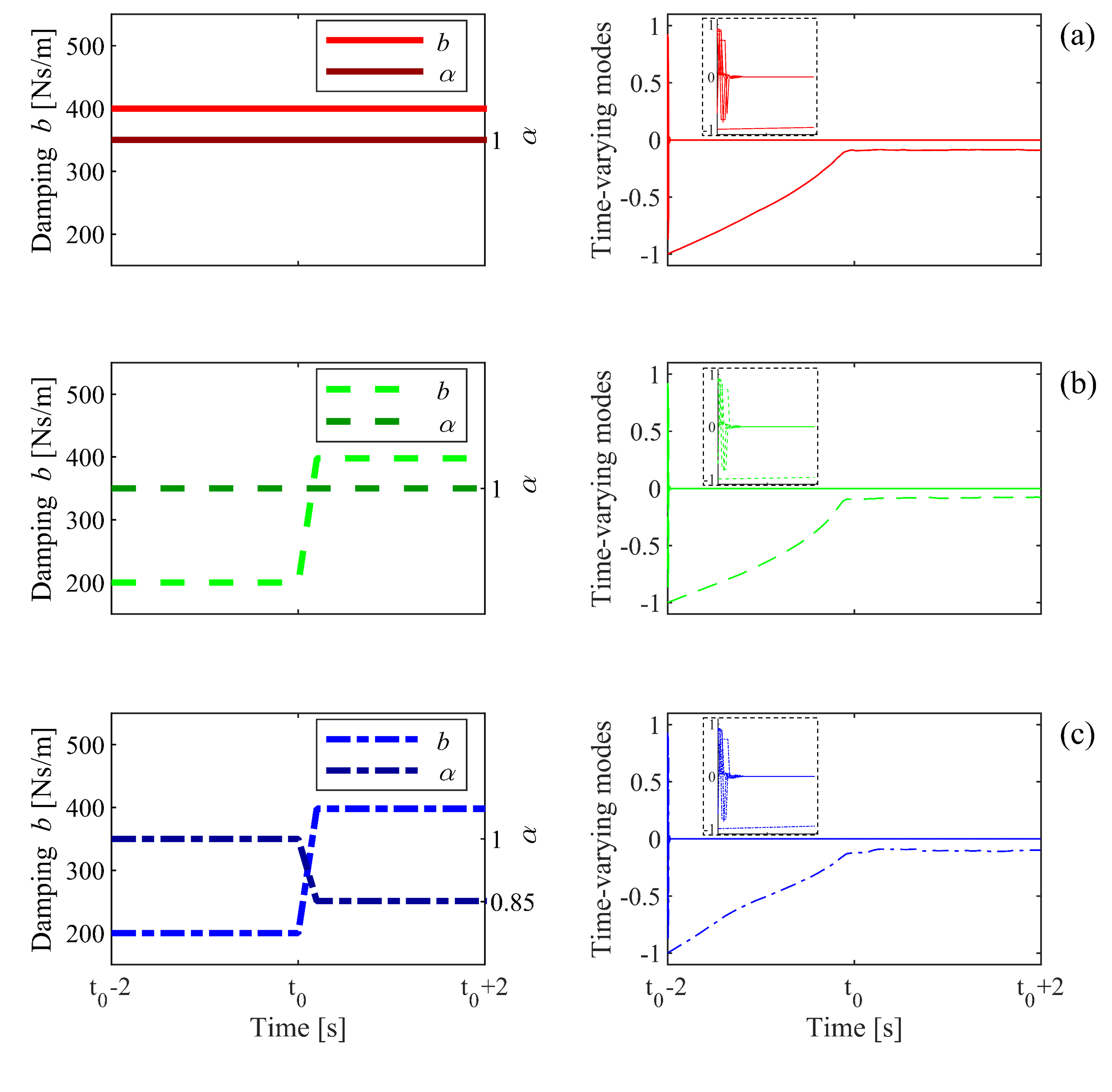}
	\centering
	\caption{Admittance damping $b$ and fractional order {\it ${\alpha}$} during the transition from {\it Driving} to {\it Contact} subtasks (left) and the corresponding first eight modes of our pHRI system (right) under three different controllers; (a) fixed IOAC, (b) adaptive IOAC, (c) adaptive FOAC. The modal analysis was performed on the experimental data for the 2 seconds before and after the contact of drill bit with the workpiece, where $t_0$ represents the time of contact. The in-line plots depict the behavior of modes for the first 0.1 seconds.
	}
	\label{fig:modes}
\end{figure}

On the other hand, since we utilize adaptive admittance controllers (C2 and C3) which are time-varying, we can inspect the modes of the system to investigate the stability. Due to time-varying nature, these modes cannot be extracted analytically, though they can be identified from the collected experimental data using the modal analysis techniques given in~\cite{Obrien2001,LIU1997,LIU1999149,Obrien1997}. We performed such analysis on the experimental data for the two seconds before and after the contact of drill bit with the workpiece. Fig.~\ref{fig:modes} shows the first eight modes of the coupled system under each controller (C1, C2, C3). As can be observed, all modes under each controller are bounded, indicating that the system is uniformly stable~\cite{Obrien2001,Bourls2011,chen1999}.

}

As a deep learning model, we opted to use ANNs. The inputs to our ANN model were (1) velocity of the end-effector, (2) interaction force, and (3) human force signals, which were measured by the sensors easily. Our analysis showed that the classification accuracy dropped to 65\% when the human force was not included as a feature and the confusion matrix suggested that the classifier differentiated \textit{Contact} phase from the others poorly. This is not surprising since the separation between the interaction force and the human force starts at the beginning of \textit{Contact} phase as shown in Fig. \ref{fig:sampletimeplot}.

\revisedS{The control parameters were adjusted by a simple linear interpolation in this study, though smoother interpolation techniques such as linear interpolation with parabolic blends or polynomial interpolation functions could be used in the implementation to prevent potential jerky behavior in adaptation.} In our approach, the successful implementation of the proposed adaptive controller heavily depends on the performance of the subtask classifier, which is affected by the following factors: \textit{(1)} \emph{delay in the response of ANN model}: The ANN model used in this study was deployed in a C++ Tensorflow environment, since our cobot utilizes C++ as the coding environment for its real-time interaction interface (called Fast Robot Interface by KUKA Inc.). The ANN model we use in our testing experiments took 2.8 ms on average to run on the Intel Xeon W-2123 CPU. This puts the effective update rate of the ANN model at 355.66 Hz, which is acceptable for realistic pHRI applications that involve contact with stiff environments such as ours. Since the ANN model runs on a separate process thread, it does not need to be updated at the same frequency of the closed-loop control system, which runs at 500 Hz in our application. \textit{(2)} \emph{delay caused by the voting buffer}: This voting buffer eliminated the instantaneous misclassifications in subtask detection as discussed earlier. Selecting a small size for this buffer increases the number of potential misclassifications while selecting a large size increases the delay in subtask classification. \textit{(3)}  \emph{delay caused by the adaptation of control parameters during subtask transitioning}: The transitioning of control parameters from one subtask to another one was handled by linear interpolation. If this transition period is too short, then the change in control parameters is abrupt and the adaptation response of the robot is not smooth for the human operator, while a long period increases the delay in transition and could cause instabilities if the transitioned subtask involves contact with a stiff environment as it is the case in our application. \revisedS{We opted for a learning-based approach rather than a rule-based one for our subtask classification problem, mainly because it would be less sensitive to new users and changes in experimental settings and environmental conditions. The subtask classification problem in our drilling task could also be solved by a rule-based approach. However, if such an approach relied on some threshold or min/max values of force and velocity signals to determine the subtasks, then those rules would need to be adjusted for new users and environmental conditions. On the other hand, defining a general set of “rules” that do not rely on some threshold or min/max values of acquired sensor signals is not easy as the task gets more complicated. As shown in Tables \ref{tab:experiments} and \ref{tab:confmat_all}, our learning-based approach worked well even when there were differences between the training and testing data sets in terms of a) users, b) experimental settings such as the drill type and the distance to workpiece, and c) environmental conditions such as the material properties of workpiece.}

One of our objectives in this study was to develop a robust ANN model for subtask classification that performs well under different conditions. Hence, we investigated if the ANN model could successfully classify the subtasks when the data for training and the testing sets came from different experiments performed with different participants under different controller and environmental conditions (inspect Table \ref{tab:experiments}). For example, the material of the workpiece used in our testing experiments was a cardboard with a thickness of 1 cm, whereas the training trials were performed with 1-cm thick plywood. Plywood is a significantly stiffer material than cardboard, and as such, not only requires higher damping values than cardboard for stability, but also requires higher human force values during drilling. However, the performance of the subtask classifier was unaffected by the stiffness of the workpiece. This proves that the ANN model successfully extracts meaningful patterns from the data, rather than memorizing only the magnitudes of the features and their thresholds. This is an important characteristic of learning-based approaches that help them transcend rule-based ones. Furthermore, the drill motor used in the testing experiment was different from the one used in the training experiment in terms of power and RPM characteristics. Again, the results showed that the classifier was not sensitive to the characteristics of the drill motor.

\section{Conclusion}
	\label{sec:Conclusion}

In the present study, an adaptive admittance controller was proposed for pHRI tasks based on time-series classification of force and velocity data by dividing the pHRI task into sequentially performed subtasks, and then assigning appropriate values for the control parameters for each subtask. Experimental results showed that an adaptive admittance control with learning-based subtask classification achieves a performance superior to that of admittance control with fixed parameters, with the added advantage of being flexible to accommodate different users and control settings, and also portable to different setups and environments. In particular, the adaptive FOAC controller (C3) resulted in significantly lower human effort for the \textit{Driving} phase and better stability characteristics for the \textit{Contact} phase.

Although the pHRI task in our study was drilling, we argue that the proposed subtask classification approach is also applicable to other small-batch manufacturing tasks such as cutting, sanding, welding, soldering, fastening, etc., which can be easily divided into three subtasks as \textit{Idle}, \textit{Driving}, and \textit{Contact}. Even though we only consider three subtasks in this study, one can also construct a larger set of subtasks for more complex pHRI tasks as well. Once such a set is constructed, the desired control parameters can be defined to meet the requirements of those subtasks. Then, an ANN can be trained to recognize the subtasks in real time, as we did it in the present study.

As a future work, a more sophisticated machine learning model can be built such that it can forecast subtask transitions before they actually happen. As a result, possible contact instabilities can be prevented during the subtask transitioning to \textit{Contact} for example. Moreover, because of the ability to anticipate a new subtask such as \textit{Contact} a few moments before it happens, much lower damping values can comfortably be selected for \textit{Driving} phase, which will further decrease human effort, leading to a higher efficiency in task performance.

\section*{CRediT authorship contribution statement}

\textbf{Berk Guler}: Methodology, Formal Analysis, Software, Experimentation, Validation, Writing – review \& editing
\textbf{Pouya P. Niaz}: Methodology, Software, Experimentation, Writing – original draft, Writing – review \& editing 
\textbf{Alireza Madani}: Methodology, Software, Experimentation, Writing – review \& editing \textbf{Yusuf Aydın}: Conceptualization, Software, Supervision, Writing – review \& editing \textbf{Cagatay Basdogan}: Conceptualization, Resources, Supervision, Funding Acquisition, Project Administration, Writing – review \& editing

\section*{Declaration of competing interest}

The authors declare that they have no known competing financial interests or personal relationships that could have appeared to influence the work reported in this paper.

\section*{Acknowledgments}

This study was supported by the Scientific and Technological Research Council of Turkey (TUBITAK) under contract number EEEAG-117E645.

\bibliography{references_v003}

\vspace{10pt}


\begin{wrapfigure}{l}{25mm} 
\includegraphics[width=1in,height=1.25in,clip,keepaspectratio]{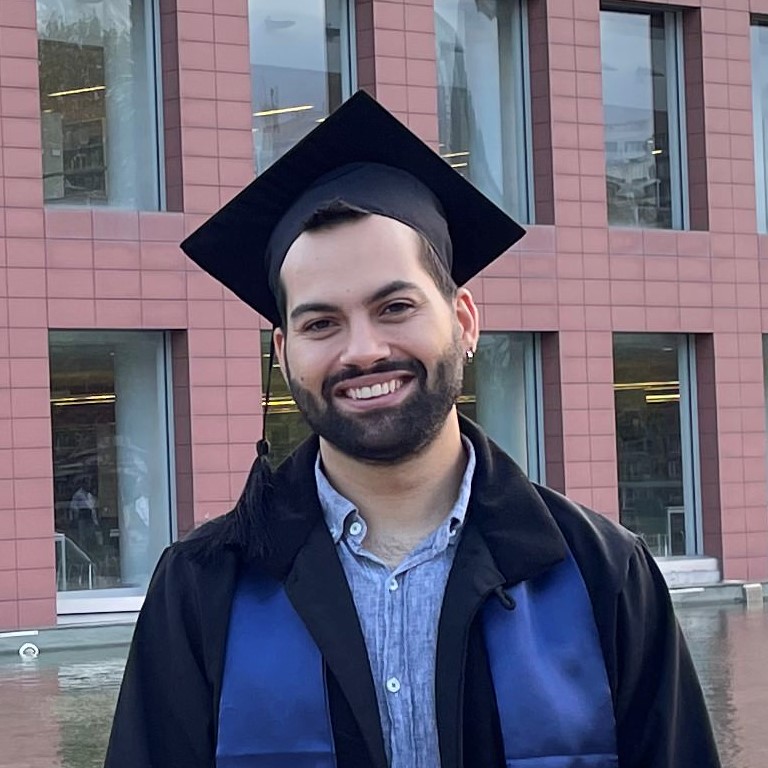}
\end{wrapfigure}\par
\textbf{Berk Guler} received his B.S. degree in Control and Automation Engineering from Istanbul Technical University in 2020. He is currently an M.S. student in Mechanical Engineering at Koç University and a research assistant at the Robotics and Mechatronics Laboratory. He is also a research fellow of the Artificial Intelligence Center at Koç University, KUIS AI Center. His research areas consist of control theory, robotics,  machine learning, mechatronics, and human-robot interaction.\par
\vspace{10pt}

\begin{wrapfigure}{l}{25mm} 
\includegraphics[width=1in,height=1.25in,clip,keepaspectratio]{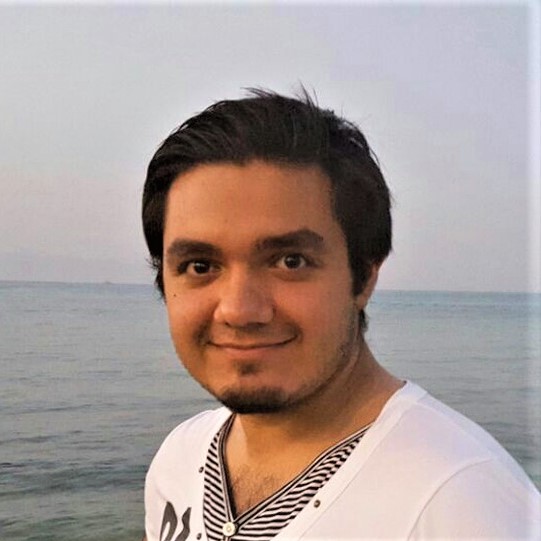}
\end{wrapfigure}\par
\textbf{Pouya P. Niaz} received his B.S. degree in Mechanical Engineering from University of Tabriz in 2016. He joined the Robotics and Mechatronics Laboratory (RML) of Koç University, Istanbul, as an M.Sc. student in Mechanical Engineering with full scholarship in 2020. He then joined the KUIS AI Center of Koç Univeristy in 2021. His research areas consist of AI, machine learning, computer vision, mechatronics, robotics, human-robot interaction, and haptics.\par

\vspace{10pt}

\begin{wrapfigure}{l}{25mm}
\centering
\includegraphics[width=1.25in,height=1.4in,clip,keepaspectratio]{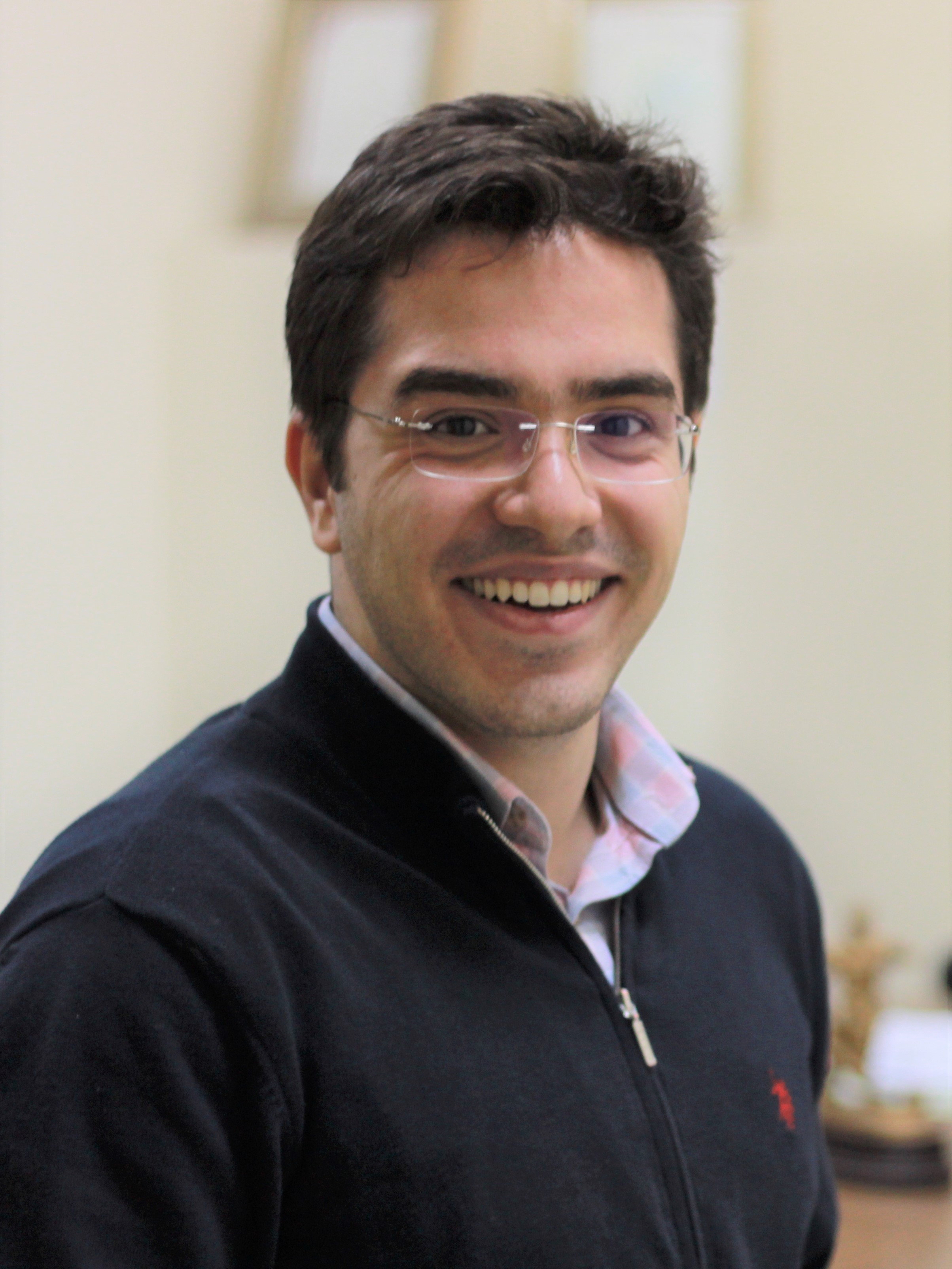}
\end{wrapfigure}\par
\textbf{Alireza Madani} received his B.Sc. in Mechanical Engineering as the 1st ranked student from K. N. Toosi University of Technology, Tehran, Iran in 2020. He is now studying M.Sc. in the same major at Koc University as research assistant in the Robotics and Mechatronics Laboratory. In 2020, he was awarded the research fellowship from KUIS AI Center. During his B.Sc. studies, he was also awarded the title of nationally distinguished engineering student. Meanwhile, he was a part-time research assistant in Advanced Robotics and Automated Systems Laboratory (ARAS-K.N.Toosi U.T.) conducting research in the area of collaborative robotic manipulation and impedance control.  His current research interests encompass dynamic systems modeling, adaptive control, deep learning for robotics, human-machine interfaces, pHRI and mechatronics.\par

\newpage

\begin{wrapfigure}{l}{25mm} 
\includegraphics[width=1in,height=1.25in,clip,keepaspectratio]{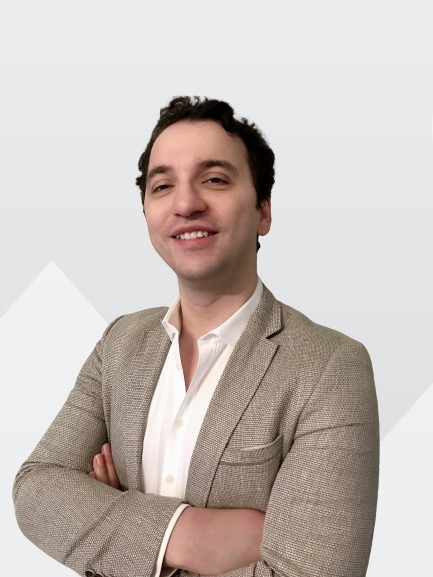}
\end{wrapfigure}\par
\textbf{Yusuf Aydin} is a faculty member in the Electrical and Electronics Engineering program at MEF University, Istanbul, Turkey. He received his B.Sc. dual degrees in Mechanical Engineering and Electrical and Electronics Engineering, M.Sc. degree in Mechanical Engineering, and then, Ph.D. degree in Mechanical Engineering from Koc University, Istanbul, in 2011, 2013, and 2019, respectively. He was awarded the prestigious TUBITAK BIDEB fellowship for his graduate studies. Before joining MEF University, he worked at Robotics and Mechatronics Laboratory at Koc University, Istanbul, as a postdoctoral research fellow. His research interests include physical human-robot interaction, haptics, robotics, control, optimization, and mechatronics.\par

\vspace{10pt}

\begin{wrapfigure}{l}{25mm} 
\includegraphics[width=1in,height=1.25in,clip,keepaspectratio]{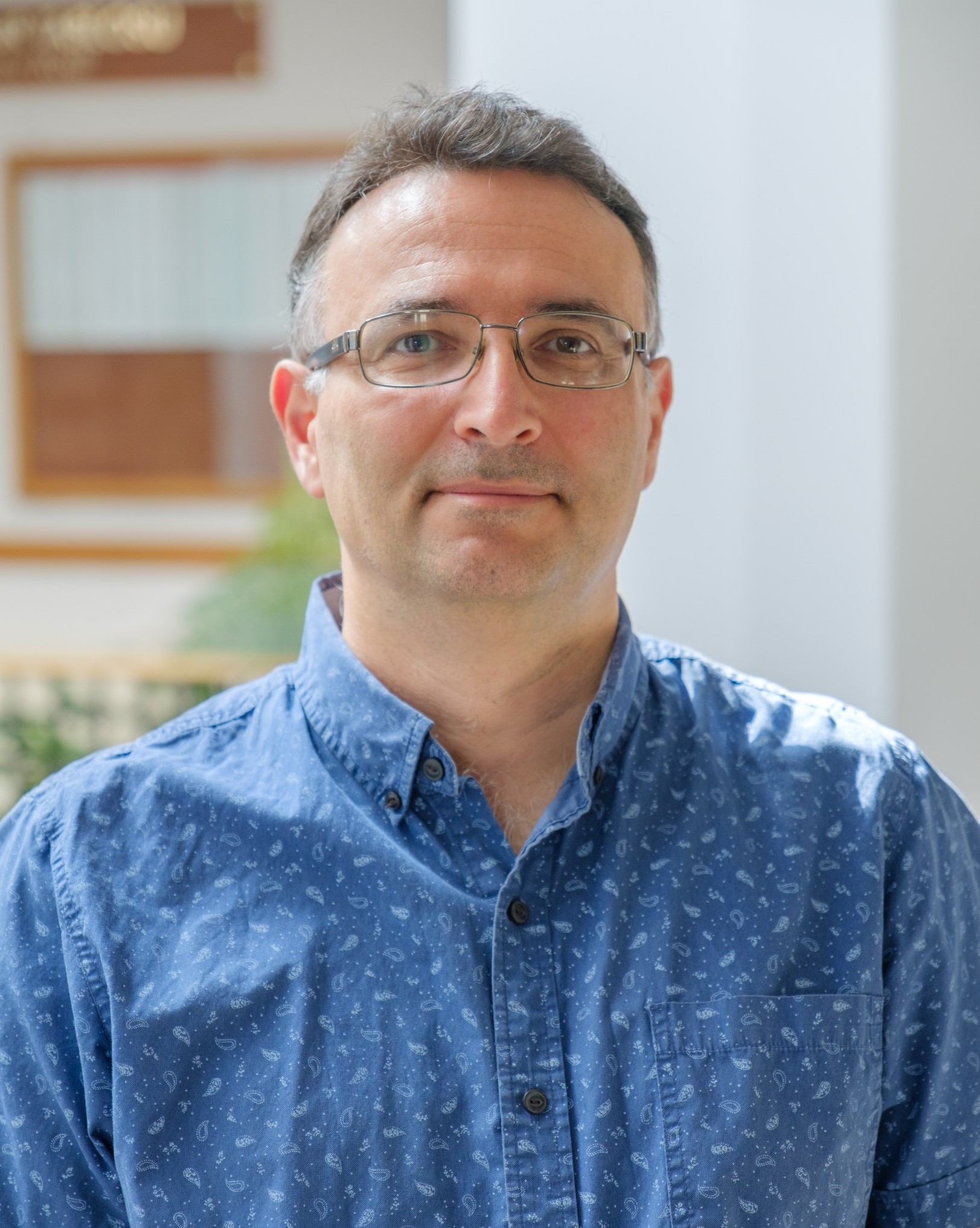}
\end{wrapfigure}\par
\textbf{Cagatay Basdogan}
received his Ph.D. degree in Mechanical Engineering from Southern Methodist University, in 1994. He is a faculty member both in the mechanical engineering and the computational sciences and engineering programs at Koc University, Istanbul, Turkey. He is the director of Robotics and Mechatronics Laboratory and affiliated with AI Center at Koc University. Before joining Koc University, he worked at NASA-JPL/Caltech, MIT, and Northwestern University Research Park. His research interests include haptic interfaces, robotics, mechatronics, biomechanics, medical simulation, computer graphics, and multi-modal virtual environments. He served on the editorial boards of the IEEE Transactions on Haptics and IEEE Transactions on Mechatronics and currently serves on the editorial boards of Presence: Teleoperators and Virtual Environments, and Computer Animation and Virtual Worlds journals. In addition to serving in programme and organizational committees of several haptics conferences, he chaired the IEEE World Haptics Conference in 2011.\par

\end{document}